\pdfoutput=1
\documentclass[11pt]{article}
\usepackage[dvipsnames]{xcolor}

\usepackage{ACL2023}

\usepackage{times}
\usepackage{latexsym}

\usepackage[T1]{fontenc}
\usepackage[utf8]{inputenc}

\usepackage{microtype}

\usepackage{inconsolata}
\usepackage{multirow}

\usepackage{graphicx}
\usepackage{colortbl}
\usepackage{booktabs}
\usepackage{CJKutf8}
\usepackage{pifont}

\definecolor{mypink}{RGB}{245, 194, 192}
\definecolor{mygray}{RGB}{242, 242, 242}
\definecolor{myyellow}{RGB}{255, 242, 204}

\title{PsyGUARD: An Automated System for Suicide Detection and Risk Assessment in Psychological Counseling}

\author{Huachuan Qiu$^{1, 2}$\quad Lizhi Ma$^{4, 5}$\thanks{\quad This work was done while she was an assistant researcher at Westlake University.} \quad Zhenzhong Lan$^{2, 3}$\thanks{\quad Corresponding author.} \\
  $^{1}$ Zhejiang University \quad $^{2}$ School of Engineering, Westlake University\\
  $^{3}$ Research Center for Industries of the Future, Westlake University\\
  $^{4}$ Zhejiang Philosophy and Social Science Laboratory\\for Research in Early Development and Childcare, Hangzhou Normal University\\
  $^{5}$ Department of Psychology, Jing Hengyi School of Education, Hangzhou Normal University\\
  \texttt{\{qiuhuachuan, lanzhenzhong\}@westlake.edu.cn}\\
}

\begin{document}
\maketitle

\begin{abstract}
As awareness of mental health issues grows, online counseling support services are becoming increasingly prevalent worldwide. Detecting whether users express suicidal ideation in text-based counseling services is crucial for identifying and prioritizing at-risk individuals. However, the lack of domain-specific systems to facilitate fine-grained suicide detection and corresponding risk assessment in online counseling poses a significant challenge for automated crisis intervention aimed at suicide prevention. In this paper, we propose PsyGUARD, an automated system for detecting suicide ideation and assessing risk in psychological counseling. To achieve this, we first develop a detailed taxonomy for detecting suicide ideation based on foundational theories. We then curate a large-scale, high-quality dataset called PsySUICIDE for suicide detection. To evaluate the capabilities of automated systems in fine-grained suicide detection, we establish a range of baselines. Subsequently, to assist automated services in providing safe, helpful, and tailored responses for further assessment, we propose to build a suite of risk assessment frameworks. Our study not only provides an insightful analysis of the effectiveness of automated risk assessment systems based on fine-grained suicide detection but also highlights their potential to improve mental health services on online counseling platforms. Code, data, and models are available at \url{https://github.com/qiuhuachuan/PsyGUARD}.
\end{abstract}

\section{Introduction}
It is well-documented that suicide is a major public health problem worldwide \citep{mann2005suicide,robinson2016social,turecki2019suicide}. Each suicide death represents a catastrophic tragedy and is reported to directly or indirectly affect many individuals, including relatives, friends, and society \citep{clark2000impact,cerel2008impact,turecki2019suicide}. Online counseling services are available in many countries, allowing for confidential and immediate help to those in need free of time and space, and therefore play a critical role in suicide prevention \citep{bialer2022detecting}, which can effectively halt user suicidal ideation \citep{maples2024loneliness}. However, the lack of domain-specific systems to enhance Chinese fine-grained suicide detection and corresponding risk assessment in online counseling poses a significant challenge for automated crisis intervention with the purpose of suicide prevention.

\begin{figure}[t!]
    \centering
    \includegraphics[width=\columnwidth]{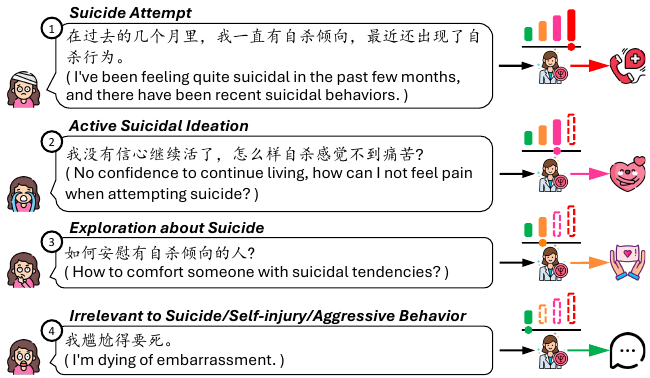}
    \caption{Examples of counselors adjusting their suicide prevention strategies based on their users' suicide actions and thoughts in real life. In this paper, the first step is to conduct fine-grained suicide detection, followed by implementing a safe, helpful, and customized approach to risk assessment.}
    \label{Fig-demo}
\end{figure}

\noindent\textbf{Motivation} Indeed, many researchers have been working on the development of automated systems for suicide detection that can be used in real-life production \citep{huang2015topic,sawhney2018computational,sawhney2022risk,sinha2019suicidal,guzman-nateras-etal-2022-event}. \textit{Despite advancements in such automated detection systems, existing studies are mainly confined to suicide detection, often ignoring fine-grained suicidal actions or thoughts and corresponding approaches for risk assessment}, as illustrated in Figure \ref{Fig-demo}. Automated suicide detection and risk assessment systems can help scale support services to reach a larger population, especially considering the increasing prevalence of online counseling and mental health support platforms. Further, by providing automated support and intervention, individuals may feel more comfortable seeking help online, thus reducing the stigma associated with mental health issues and suicide prevention \citep{robinson2016social}. Empirical evidence \citep{nie2024llm,maples2024loneliness} indicates that individuals are willing to interact with human or AI counselors, with many having disclosed their suicidal thoughts, plans, and actions, underscoring the importance of automated systems for suicide detection and risk assessment. The deficiencies in existing systems significantly limit automated systems' capability to ensure safe, helpful, customized services in providing mental health support, which motivates us to carry out the work presented in this paper.

\noindent\textbf{Challenges} \textit{The lack of fine-grained suicide detection datasets in psychological counseling is a major challenge.} Currently, numerous studies have made significant progress in detecting suicidal ideation, but they primarily focus on social media platforms \citep{huang2015topic,cao2019latent,sawhney2018computational,sinha2019suicidal,gaur2019knowledge,guzman-nateras-etal-2022-event} rather than on counseling conversations. Therefore, using such datasets directly for risk detection in online counseling may not be suitable due to a gap in user expressions, such as emojis, URLs, images, or special marks. Additionally, challenges posed by datasets collected from electronic health records \citep{pratap2022scan} or mental health records also include gaps in data format. Furthermore, most studies that primarily focus on binary suicidal ideation detection \citep{huang2015topic,cao2019latent,sawhney2018computational,sinha2019suicidal} face challenges in considering the granularity of suicide ideation categories in the real world.

\textit{The lack of a comprehensive suite of risk assessments for corresponding suicide categories is another challenge.} In addition to users mentioning that they have attempted suicide, simply identifying fine-grained suicide categories is not sufficient to conclude whether a user will actually commit suicide. Risk assessment can directly guide how to intervene in a crisis situation. Therefore, suicide classification is the initial step in suicide prevention, and further risk assessment is required, which is largely overlooked by current studies.

\noindent\textbf{Our Approach} In this paper, to our knowledge, we are the first to propose to study an automated system for suicide detection and risk assessment in psychological counseling. Our paper is organized into five main parts. Section 2 (§\ref{related-work}) describes the existing works related to ours. Section 3 (§\ref{taxonomy}) demonstrates the detailed process of taxonomy construction. Section 4 (§\ref{data-collection}) elaborates on rigorous data collection. Section 5 (§\ref{baselines}) constructs extensive baselines, and Section 6 (§\ref{risk-assessment}) provides a simple yet effective framework for risk assessment prior to crisis intervention.

\begin{table*}[ht]
\centering
\scalebox{0.55}{
\begin{tabular}{lllllllllll}
\toprule
\textbf{Dataset}                   & \textbf{Source }                                                                                & \textbf{\# Classes} & \textbf{Size}  & \textbf{Balance}                                                  & \textbf{Open-source}    & \textbf{Language} & \textbf{Level} & \textbf{Actions or Thoughts} & \textbf{Multi-label} & \textbf{Annotators} \\ \hline
\citet{huang2015topic}          & Weibo                                                                                  & 2          & 7314  & \begin{tabular}[c]{@{}l@{}}9.08\%\\ (664)\end{tabular}   & \ding{56}               & Chinese  & \ding{56}     & \ding{56}     & \ding{56} & Experts           \\ \hline
\citet{cao2019latent}          & Weibo                                                                                & 2          & 744031 & \begin{tabular}[c]{@{}l@{}}34.00\%\\ (252901)\end{tabular} & \ding{56}               & Chinese  & \ding{56}     & \ding{56}     & \ding{56} & Experts          \\ \hline
\citet{sawhney-etal-2018-computational}        & Twitter                                                                                & 2          & 5213  & \begin{tabular}[c]{@{}l@{}}15.76\%\\ (822)\end{tabular}  & \ding{56}               & English  & \ding{56}     & \ding{56}     & \ding{56} & Experts          \\ \hline
\citet{sinha2019suicidal}          & Twitter                                                                                & 2          & 34306 & \begin{tabular}[c]{@{}l@{}}11.61\%\\ (3984)\end{tabular} & \ding{56}               & English  & \ding{56}     & \ding{56}     & \ding{56} & Experts          \\ \hline
\citet{gaur2019knowledge}          & Reddit                                                                                & 5          & 500 & \begin{tabular}[c]{@{}l@{}}58.6\%\\ (293)\end{tabular} & \ding{52}               & English  & \ding{52}     & \ding{56}     & \ding{56}  & Experts         \\ \hline

\citet{guzman-nateras-etal-2022-event} & Reddit                                                                                 & 7          & 37068 & \begin{tabular}[c]{@{}l@{}}20.85\%\\ (7729)\end{tabular} &\begin{tabular}[c]{@{}l@{}}\ding{52}\end{tabular} & English  & \ding{56}     & \ding{56}     & \ding{52} & Experts          \\ \hline
\textbf{PsySUICIDE (Ours)}         & \begin{tabular}[c]{@{}l@{}}Zhihu, Weibo, Yixinli,\\ Open-source dialogues\end{tabular} & 11         & 15010 & \begin{tabular}[c]{@{}l@{}}20.68\%\\ (3104)\end{tabular} & \ding{52}               & Chinese  & \ding{52}     & \ding{52}     & \ding{52}  & Experts       \\ \bottomrule
\end{tabular}
}
\caption{Comparison of suicidal ideation detection datasets. Our dataset provides a well-balanced representation of specific categories: (1) Suicidal action or ideation (3104/15010, accounting for 20.68\%); (2) Suicidal action or ideation, self-injury, aggressive behavior, and exploration about suicide (4256/15010, accounting for 28.35\%).}
\label{Tab-comparison}
\end{table*}

\paragraph{Our Contributions} We believe our work offers a new perspective on building an automated system for suicide detection and risk assessment in psychological counseling within the research community. Our contributions can be summarized as follows:

\begin{itemize}
    \item We construct PsyGUARD, an automated system for suicide detection and risk assessment, to ensure safe, helpful, customized services in text-based counseling conversations. To achieve this, we develop a novel fine-grained taxonomy (§\ref{taxonomy}) for crisis situations, which categorizes the risk levels based on suicidal actions or thoughts, self-harm or harming others, and being abused.
    \item We build the PsyGUARD dataset, a large-scale, high-quality, and fine-grained suicidal ideation detection corpus (§\ref{data-collection}). This dataset is created through a rigorous collection process, including raw data collection, development of annotation platforms, initial annotator training, iterative human annotation, disagreement adjudication, and quality control.
    \item To understand the capabilities of automated systems in suicide risk detection, we establish various baselines (§\ref{baselines}) using our dataset for comparison. These baselines include LLM zero-shot, LLM few-shot, fine-tuning pre-trained models, and fine-tuning LLM used for predicting suicidal ideation of users' content.
    \item To assist automated services in providing safe, helpful, and customized responses during risk assessment, we propose to build a risk assessment framework (§\ref{risk-assessment}) for users during online text-based counseling.
\end{itemize}

Next, we will briefly describe the existing works related to ours.

\section{Related Work}
\label{related-work}

\subsection{Taxonomy for Suicide Risk}
Existing suicide risk annotations are mainly based on the guidelines of the Columbia Suicide Severity Rating Scale (C-SSRS) \citep{posner2008columbia,posner2011columbia}, which is an authoritative questionnaire used by psychiatrists to assess the severity of suicide risk. Each C-SSRS severity class comprises a set of questions that conceptually characterize the respective category. The responses to these questions across the C-SSRS classes determine the risk of suicidality for an individual \citep{gomes2018ethics,mccall2021examining,orr2022ethical}. Additionally, there is another commonly used taxonomy \citep{shing2018expert,zirikly2019clpsych} for suicide annotation, which includes four levels: no risk, low risk, moderate risk, and severe risk. Compared to the C-SSRS, this taxonomy may have varying degrees of subjectivity. Furthermore, a more easily understandable taxonomy \citep{sawhney2018computational} is the binary classification system, which categorizes individuals as either having present or absent suicidal intent. Clearly, the existing taxonomies are either too simplistic or too complex, and they do not fully meet the requirements of our research purpose.

\subsection{Suicide Risk Detection}

\subsubsection{Datasets for Suicide Risk Detection}
We present several typical datasets used for suicide detection in Table \ref{Tab-comparison}. Various works have been recently proposed with the objective of automating the detection of user content expressing suicidal ideation posted on social media platforms \citep{huang2015topic,cao2019latent,sawhney2018computational,sawhney2022risk,sinha2019suicidal,guzman-nateras-etal-2022-event} and electronic health records \citep{pratap2022scan}. Further, some researchers focus on electronic health records \citep{guzman-nateras-etal-2022-event,rawat-etal-2022-scan} to detect clinical health issues.

\subsubsection{Methods for Suicide Risk Detection}
In short, the best available performance for suicidal ideation detection still relies heavily on pre-trained models. However, in order to improve performance, researchers have added a variety of strategies to enhance the model's ability to classify \citep{rawat2022parameter,ghosh2022persona,sawhney2022risk}. Basically, most of the research focused on conventional machine learning methods \citep{tyagi-etal-2023-trigger} and fine-tuning pre-trained models \citep{sawhney-etal-2020-time,shing-etal-2020-prioritization,sawhney-etal-2022-risk}. In the era of large language models, one work \citep{ghanadian-etal-2023-chatgpt} conducted a quantitative analysis of the open-source suicide intent classification dataset using ChatGPT, evaluating methods including zero-shot and few-shot paradigms. The experimental results have much room for improvement, but it is a crucial attempt and exploration of using large models for suicide detection.

\begin{figure*}[t!]
    \centering
    \includegraphics[width=\textwidth]{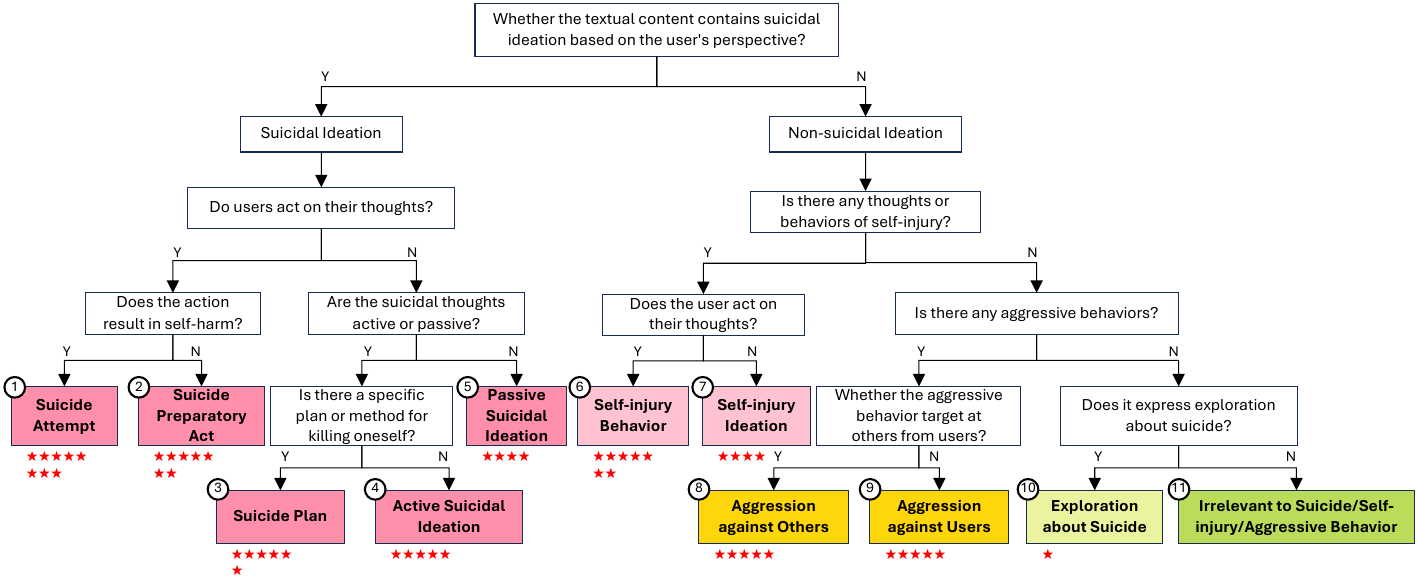}
    \caption{Our proposed taxonomy of suicidal ideation. The higher the number of stars, the higher the risk. See Figure \ref{Fig-taxonomy-zh} in the Appendix for the corresponding Chinese version.}
    \label{Fig-taxonomy}
\end{figure*}

\section{Taxonomy Construction}
\label{taxonomy}
To build an automated system for suicide detection and risk assessment in psychological counseling, we first propose to develop a novel taxonomy for categorizing the level of suicide based on suicide actions and thoughts. In collaboration with experts\footnote{One is a Ph.D. in psychology and holds a State-Certificated Class 3 Psycho-counselor with four years of experience in counseling. Another individual is a State-Certificated Class 3 Psycho-counselor with a master's degree. The third person is a doctoral student majoring in computer science and is the first author of this paper.} in psychological counseling, we have adapted and refined existing suicidal taxonomies, such as C-SSRS \citep{posner2008columbia,posner2011columbia}, dichotomy suicide \citep{sawhney2018computational}, suicide behaviors \citep{nock2008suicide,crosby1999incidence,schreiber2010suicidal}, self-injury behaviors \citep{nock2010self}, and aggressive behavior \citep{stanford2003characterizing,grigg2010cyber}, to suit the context of online text-based counseling conversations. Based on the solid theories and preliminary analysis of the real-life corpus, we elaborately construct the suicide taxonomy, following the consensual qualitative research method \citep{nock2008suicide,bridge2006adolescent}.

\paragraph{Category Definitions} Our proposed taxonomy for suicide ideation detection, which consists of a total of 11 fine-grained categories, is presented in Figure \ref{Fig-taxonomy}. Based on the user's perspective, whether the textual content contains suicidal ideation can be divided into two key types, as illustrated by dichotomy suicide, including suicidal ideation and non-suicidal ideation. For details about the definition of each category, see Appendix \ref{App-definitions}.

\section{Data Collection}
\label{data-collection}
\begin{table*}[t!]
    \centering
    \includegraphics[width=\textwidth]{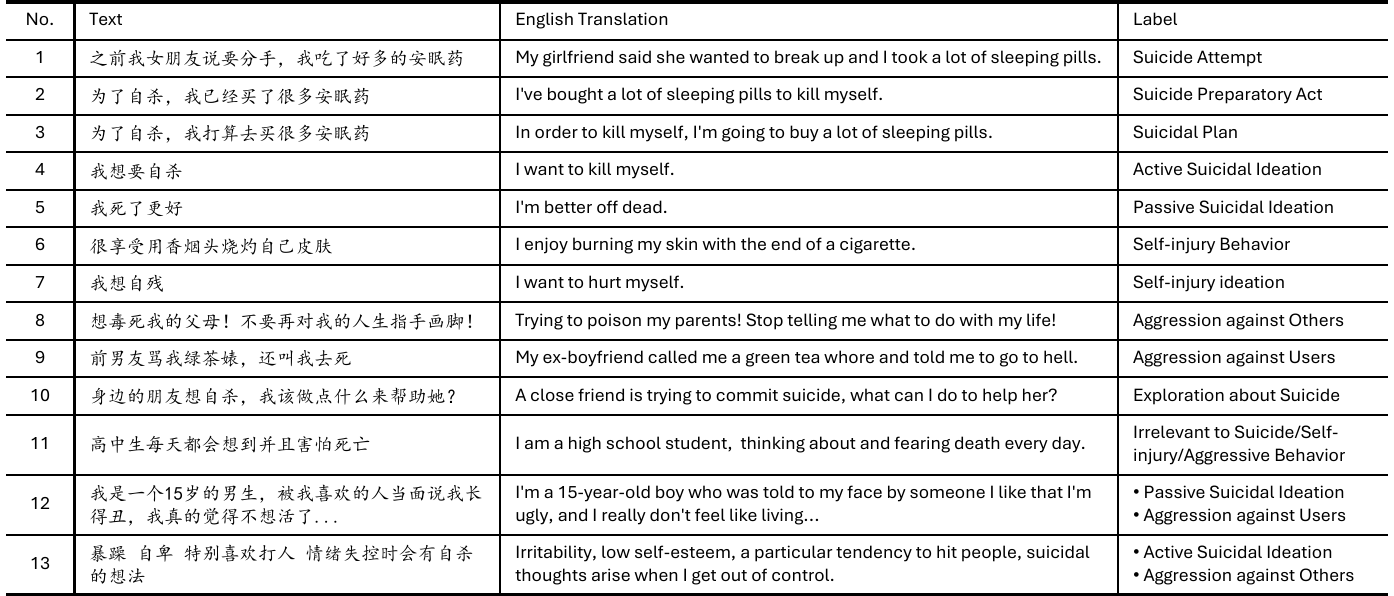}
    \caption{Examples that are cherry-picked from the PsySUICIDE dataset.}
    \label{Tab-examples}
\end{table*}

To validate the feasibility of our proposed fine-grained taxonomy in real-world settings and address the limitations that current datasets are typically collected from a single source, as well as to automate risk assessment further, we collect a large-scale user input corpus from several media platforms, including Weibo\footnote{https://www.weibo.com}, Zhihu\footnote{https://www.zhihu.com}, and Yixinli\footnote{https://www.xinli001.com}, and open-source dialogue datasets. Through a rigorous process of data annotation, the PsySUICIDE dataset is a \textit{high-quality} and \textit{diverse} corpus used for researching automated suicidal detection and further risk assessment. Some examples cherry-picked from the PsySUICIDE dataset are presented in Table \ref{Tab-examples}.

Next, we will describe the process of constructing PsySUICIDE, including raw data collection, development of an annotation platform, initial annotator training, iterative human annotation, disagreement adjudication, and quality control. Finally, we will present the statistics of the PsySUICIDE dataset, as illustrated in Table \ref{Tab-data-statistics}.

\subsection{Raw Data Collection}
To cover both frequent explicit or implicit suicidal ideation, such as explicit suicidality and queries about committing suicide, as well as infrequent non-suicidal ideation, such as reports of suicide and exploration about suicide, we have selected the raw data for our dataset from two publicly accessible sources: social media platforms and open-source mental health dialogues. Furthermore, to ensure that the data is suitable for conversational scenarios, we primarily collect our data from open-source dialogues that focus on mental health counseling.

To ensure that there is no data duplication in the unlabeled dataset, we initially performed duplication filtering. To protect user privacy, we then asked annotators to conduct a data anonymization process, removing information related to user identification (e.g., names and addresses) before the data annotation.

We collect 3800 user posts from social media platforms. Furthermore, we collect 11000 user utterances from open-source mental health dialogues, with 4000, 3000, and 4000 user utterances from SmileChat \citep{qiu2023smile}, real-life counselor-client dialogues \citep{li2023understanding}, and real-life human-machine dialogues collected in the wild \citep{qiu2023smile}, respectively. For details of data sources, see Appendix \ref{App-details-of-data-sources}.

\begin{table}[t]
\centering
\scalebox{0.64}{
\begin{tabular}{ll}
\toprule
\textbf{Data Category} & \textbf{\# Instances} \\ \hline
single-label & 14594 \\
multi-label & 206 \\ \hline
Total & 14800 \\ \hline\hline
\toprule
\textbf{Label} & \textbf{\# Number} \\ \hline
Suicide Attempt & 118 \\
Suicidal Preparatory Act & 22 \\
Suicidal Plan & 155 \\
Active Suicidal Ideation & 1430 \\
Passive Suicidal Ideation & 1379 \\
Self-injury Behavior & 160 \\
Self-injury Ideation & 48 \\
Aggression against Others & 315 \\
Aggression against Users & 260 \\
Exploration about Suicide & 369 \\
Irrelevant to Suicide/Self-injury/Aggressive Behavior & 10754 \\ \hline
Total & 15010 \\ \bottomrule
\end{tabular}
}
\caption{Data statistics of the PsySUICIDE dataset.}
\label{Tab-data-statistics}
\end{table}

\subsection{Annotation Platform Development}
We present our annotation platform based on our proposed taxonomy, which consists of at least three tasks and at most four tasks, as illustrated in Figure \ref{Fig-annotation-interface} in Appendix \ref{App-annotation-framework}. We will release this annotation platform along with our code, dataset, and model.

\subsection{Initial Annotator Training}
Three annotators are undergraduate fourth-year students majoring in psychology, with two of them being male and one being female. We provide our taxonomy (Figure \ref{Fig-taxonomy}) and annotation guidelines along with concrete examples (Table \ref{Tab-guidelines-zh} in Appendix \ref{Sec-annotation-guidelines}) for three annotators. Prior to data annotation, we require three annotators to understand our taxonomy and annotation guidelines. Any questions they have about their understanding have been resolved by our experts, thus ensuring that we have reached an agreement before labeling.

\paragraph{Trial-and-Error Annotation} To validate the feasibility of the initial taxonomy and reduce its obscure points, we propose to adopt a trial-and-error annotation paradigm to annotate three batches of data, comprising 200, 300, and 300 instances, respectively. Fleiss' kappa \citep{fleiss1981measurement} is used to measure the inter-rater agreement, and all values (0.555, 0.511, and 0.565) fall within moderate agreement with $0.5\le \kappa \le0.6$. After three batch annotations, we discuss the cases in which one annotator assigns a different label. Consequently, we improve our taxonomy based on the real-life corpus. Through trial-and-error annotation, in cooperation with our experts and three annotators majoring in psychology, we update the taxonomy again.

\subsection{Iterative Human Annotation}
We adopt two-stage data annotation, including mini-batch iterative annotation and large-scale iterative annotation. Each batch contains a certain amount of content produced by users, and each unlabeled instance is assigned to three annotators for independent annotation using our annotation platform.

\paragraph{Mini-batch Iterative Annotation} To validate the completeness of our taxonomy, we assign five batches, each containing 100 instances. Fleiss' kappa is used to measure the inter-rater agreement, and all values (0.739, 0.74, 0.784, 0.785, and 0.816) fall within substantial agreement or even almost perfect with $0.7\le \kappa \le0.9$, which demonstrates that our taxonomy is complete enough.

\paragraph{Large-scale Iterative Annotation}
We assign 27 batches of data for large-scale iterative annotation, each containing 500 instances. Fortunately, Fleiss' kappa value in each batch is consistently higher than 0.7, demonstrating that the annotated data is of high quality with substantial agreement.

\subsection{Disagreement Adjudication}
In any batch of data annotation, we first use majority voting to resolve label disagreements. When all three labels are distinct, the three annotators must discuss any inconsistent instances that have not been assigned the same label. Three annotators are required to discuss the final label for any instance assigned a distinct label for disagreement adjudication.

It is worth noting that some instances have multiple labels. Therefore, we require all annotators to tick the option if such an instance has multiple labels. During disagreement adjudication, we also require them to discuss such instances and assign correct labels in such cases.

\subsection{Quality Control}
To ensure high-quality annotation, we set a rigorous annotation standard: If the Fleiss' kappa value is lower than 0.6, the entire batch is rejected and returned to the annotators for revision until the Fleiss' kappa value exceeds 0.6. There are a total of 27 batches of data in the process of large-scale iterative human annotation. Upon completing a batch, we calculate the Fleiss' kappa value and conduct statistics on inconsistent instances. We report all Fleiss' kappa values during large-scale iterative annotation in Table \ref{Tab-quality-control} in Appendix \ref{Sec-quality-control}.

\subsection{Data Statistics}
We present the data statistics of the PsySUICIDE dataset in Table \ref{Tab-data-statistics}. For details of length distribution, refer to Table \ref{Tab-CLI}. There are 14800 instances in our dataset, with 14594 instances having a single label and 206 instances having multiple labels. Only 22 instances contain the label of suicidal preparatory act, demonstrating that in real-life chatting scenarios, users often do not disclose their actions in preparation for suicide. The average Chinese character length per user utterance is 30.

For training, validation, and test sets, each set is generated by stratified random sampling \citep{scikit-learn} from the annotated dataset to maintain consistency in data distribution, with a partition ratio of 8:1:1. Specifically, in terms of single-label instances, we first group them by labels and split them with a stratified random sampling strategy. For simplicity, we directly split multi-label instances using a random sampling strategy.

\section{Automated System for Suicide Detection}
\label{baselines}

\begin{table*}[t!]
\centering
\scalebox{0.7}{
\begin{tabular}{lccccccc}
\toprule
\textbf{Model} & \textbf{Accuracy} & \textbf{Micro P.} & \textbf{Micro R.} & \textbf{Micro F1.} & \textbf{Macro P.} & \textbf{Macro R.} & \textbf{Macro F1.} \\ \hline
\textsc{ChatGLM2-6B-zero-shot} & $1.17_{0.16}$ & $6.78_{0.13}$ & $34.77_{1.31}$ & $11.35_{0.24}$ & $9.29_{0.11}$ & $41.22_{0.29}$ & $7.96_{0.05}$ \\ \hline
\textsc{ChatGLM2-6B-few-shot} & $0.56_{0.14}$ & $9.30_{0.05}$ & $76.80_{1.00}$ & $16.59_{0.05}$ & $9.35_{0.03}$ & $93.15_{1.27}$ & $12.29_{0.03}$ \\ \hline
\textsc{Qwen1.5-1.8B-Chat-zero-shot} & $4.51_{0.18}$ & $5.27_{0.25}$ & $5.97_{0.30}$ & $5.60_{0.27}$ &$10.32_{0.65}$ & $12.96_{0.95}$& $3.31_{0.86}$ \\ \hline
\textsc{Qwen1.5-1.8B-Chat-few-shot} & $1.26_{0.26}$ & $8.30_{0.14}$ & $51.49_{1.64}$ & $14.29_{0.28}$ &$9.26_{0.11}$ & $58.26_{6.73}$& $9.97_{0.22}$ \\ \hline
\textsc{Qwen1.5-4B-Chat-zero-shot} & $22.45_{1.11}$ & $24.26_{1.33}$ & $23.38_{1.07}$ & $23.81_{1.19}$ & $18.95_{0.47}$ & $29.99_{0.84}$ & $15.97_{0.58}$ \\ \hline
\textsc{Qwen1.5-4B-Chat-few-shot} & $21.35_{0.47}$ & $19.15_{0.19}$ & $28.42_{0.21}$ & $22.89_{0.16}$ & $13.78_{0.22}$ & $35.57_{3.30}$ & $12.20_{0.43}$ \\ \hline
\textsc{Qwen1.5-7B-Chat-zero-shot} & $60.38_{0.19}$ & $61.76_{0.31}$ & $60.65_{0.23}$ & $61.20_{0.27}$ & $25.56_{0.15}$ & $38.41_{1.21}$ & $27.57_{0.28}$ \\ \hline
\textsc{Qwen1.5-7B-Chat-few-shot} & $63.48_{0.41}$ & $62.79_{0.68}$ & $66.20_{0.54}$ & $64.45_{0.60}$ & $28.41_{1.36}$ & $46.83_{4.15}$ & $28.31_{2.11}$ \\ \hline
\textsc{Qwen1.5-14B-Chat-zero-shot} & $31.27_{0.10}$ & $31.89_{0.03}$ & $32.03_{0.04}$ & $31.96_{1.18}$ & $37.58_{1.18}$ & $40.35_{0.87}$ & $27.48_{0.77}$ \\ \hline
\textsc{Qwen1.5-14B-Chat-few-shot} & $69.18_{0.71}$ & $67.78_{0.69}$ & $71.11_{0.63}$ & $69.41_{0.66}$ & $34.72_{1.23}$ & $50.48_{1.83}$ & $36.38_{1.35}$ \\ \hline
\textsc{Qwen1.5-32B-Chat-zero-shot} & $67.83_{0.25}$ & $68.30_{0.29}$ & $67.86_{0.28}$ & $68.08_{0.28}$ &$43.63_{0.39}$ & $48.71_{0.78}$& $38.01_{0.45}$ \\ \hline
\textsc{Qwen1.5-32B-Chat-few-shot} & $78.47_{0.14}$ & $77.41_{0.48}$ & $80.03_{0.37}$ & $78.70_{0.42}$ &$49.73_{1.83}$ & $56.78_{0.56}$& $48.63_{1.27}$ \\ \hline
\textsc{Qwen1.5-72B-Chat-zero-shot} & $61.64_{0.33}$ & $61.94_{0.48}$ & $62.64_{0.23}$ & $62.29_{0.35}$ &$36.11_{0.41}$ & $54.62_{0.70}$& $38.61_{0.13}$ \\ \hline
\textsc{Qwen1.5-72B-Chat-few-shot} & $69.43_{0.47}$ & $69.43_{0.46}$ & $71.42_{0.50}$ & $70.41_{0.48}$ &$39.66_{1.09}$ & $55.79_{1.15}$& $43.07_{1.16}$ \\ \hline
\textsc{GPT-3.5-zero-shot} & $61.19_{0.81}$ & $61.95_{0.76}$ & $61.34_{0.83}$ & $61.64_{0.79}$ &$32.73_{1.67}$ & $46.90_{2.34}$& $34.50_{1.74}$\\ \hline
\textsc{GPT-3.5-few-shot} & $71.13_{0.35}$ & $70.49_{0.45}$ & $74.23_{0.64}$ & $72.31_{0.53}$ &$38.99_{2.17}$ & $52.52_{1.57}$& $41.97_{1.41}$\\ \hline
\textsc{GPT-4-preview-zero-shot} & $82.72_{0.21}$ & $83.59_{0.13}$ & $82.73_{0.10}$ & $83.16_{0.11}$ &$54.10_{1.82}$ & $54.31_{1.56}$& $49.32_{1.55}$\\ \hline
\textsc{GPT-4-preview-few-shot} & $81.73_{0.35}$ & $81.86_{0.41}$ & $82.66_{0.21}$ & $82.26_{0.31}$ &$48.79_{1.49}$ & $61.62_{1.28}$& $51.64_{0.81}$\\ \hline
\textsc{GPT-4-zero-shot} & $74.77_{0.37}$ & $75.19_{0.44}$ & $76.20_{0.47}$ & $75.69_{0.45}$ &$43.13_{0.53}$ & $67.97_{1.67}$& $48.95_{0.69}$\\ \hline
\textsc{GPT-4-few-shot} & $71.87_{0.35}$ & $71.70_{0.30}$ & $78.79_{0.44}$ & $75.08_{0.33}$ &$42.42_{0.34}$ & $71.48_{1.61}$& $49.30_{0.63}$\\ \hline
\textsc{BERT-base} & $90.77_{0.37}$ & $92.39_{0.37}$ & $91.64_{0.30}$ & $92.01_{0.31}$ & $70.55_{3.46}$ & $62.70_{2.03}$& $64.89_{2.22}$\\ \hline
\textsc{RoBERTa-large} & \colorbox{YellowOrange}{\textbf{91.69$_{0.39}$}} & \colorbox{Lavender}{\textbf{92.94$_{0.39}$}} & \colorbox{Lavender}{\textbf{92.59$_{0.43}$}} & \colorbox{Lavender}{\textbf{92.77$_{0.40}$}} & \colorbox{Lavender}{\textbf{73.43$_{1.74}$}} & \colorbox{YellowOrange}{\textbf{68.03$_{1.88}$}} & \colorbox{YellowOrange}{\textbf{69.76$_{1.48}$}}\\ \hline
\multirow{3}{*}{\textsc{ChatGLM2-6B-lora}} & $91.83_{0.22}$ & $92.27_{0.20}$ & $92.37_{0.20}$ & $92.32_{0.20}$ &\colorbox{YellowOrange}{\textbf{72.68$_{0.76}$}} & \colorbox{Lavender}{\textbf{72.83$_{1.09}$}}& \colorbox{Lavender}{\textbf{72.19$_{0.35}$}}\\
 & $91.69_{0.14}$ & $92.05_{0.25}$ & $92.19_{0.14}$ & $92.12_{0.20}$ &$71.97_{1.17}$ & $71.74_{1.26}$& $70.61_{0.54}$\\
 & \colorbox{Lavender}{\textbf{91.99$_{0.24}$}} & \colorbox{YellowOrange}{\textbf{92.38$_{0.23}$}} & \colorbox{YellowOrange}{\textbf{92.52$_{0.28}$}} & \colorbox{YellowOrange}{\textbf{92.45$_{0.25}$}} &$72.32_{2.64}$ & $71.00_{1.62}$& $70.63_{1.59}$\\ \bottomrule
\end{tabular}
}
\caption{Evaluation results for fine-grained classification on the test set, with the \colorbox{Lavender}{best} and \colorbox{YellowOrange}{second best} are highlighted with cell color. The results denote the mean and standard deviation (subscript) of accuracy (Acc.), precision (P.), recall (R.), and F1-score (F1.). Regarding LoRA tuning, we only select the best result for comparisons. In each row of ChatGLM2-6B-LoRA models, corresponding seeds are 42, 43, and 44 in order.}
\label{tab:results}
\end{table*}

We conduct our experiments using pre-trained language models (LMs) and large language models (LLMs). All experiments in this paper are performed on NVIDIA A100 8 $\times$ 80G GPUs.

\subsection{Prompt-based Paradigm for LLMs}
\paragraph{LLMs} We prompt several popular LLMs to elicit textual labels via instructions free of fine-tuning, including zero- and few-shot settings. In this paper we propose to evaluate several popular open-source LLMs, such as ChatGLM2-6B \citep{zeng2022glm}, Qwen1.5-1.8B-Chat, Qwen1.5-4B-Chat, Qwen1.5-7B-Chat, Qwen1.5-14B-Chat, Qwen1.5-32B-Chat and Qwen1.5-72B-Chat \citep{qwen}. Additionally, we also evaluate three popular closed-source LLMs \citep{openai2024gpt4}, such as GPT-3.5 Turbo\footnote{The model we use is gpt-3.5-turbo-0125, with training data up to Sep 2021.}, GPT-4-preview\footnote{The model we use is gpt-4-1106-preview, with training data up to Apr 2023.} and GPT-4\footnote{The model we use is gpt-4-0613, with training data up to Sep 2021.}, where GPT-4-preview and GPT-4 are state-of-the-art models acknowledged by researchers.

\paragraph{Experimental Setup} Due to the generation diversity in LLMs, we propose to prompt LLMs to generate exact labels given an instruction and an unlabeled instance three times. Based on official recommendations, we set the \verb|temperature| and \verb|top_p| to 0.8 and 0.8 for ChatGLM2-6B, 0.7 and 0.8 for the Qwen series, and 1.0 and 1.0 for the OpenAI GPT series, respectively.

\paragraph{Prompting Method}
The zero-shot prompting template is presented in Figure \ref{Fig-prompt-template-zh-zero-shot}. The few-shot prompting template is provided in Figure \ref{Fig-prompt-template-zh-few-shot}. The in-context examples are fixed and selected from Table \ref{Tab-examples}, including 13 instances.

\subsection{Fine-tuning Pre-trained LMs}
\paragraph{Pre-trained LMs} We apply two pre-trained models, BERT \citep{devlin2018bert} and RoBERTa \citep{liu2019roberta}, which are popular language models with only an encoder architecture used widely in various tasks in natural language processing, to train a text classification model. In this paper, we fine-tune the entire \textsc{BERT-base}\footnote{The model we use is google-bert/bert-base-chinese.} and \textsc{RoBERTa-large}\footnote{The model we use is hfl/chinese-roberta-wwm-ext-large.} models.

\paragraph{Experimental Setup} Considering that the hyper-parameters for fine-tuning the pre-training model appear in numerous papers and are not the core part of this paper, we place this section in Appendix \ref{App-setup}.

\subsection{LLM Parameter-efficient Fine-tuning}
Like fine-tuning the BERT and RoBERTa models, we select one of the most widely used open-source models, ChatGLM2-6B to conduct parameter-efficient fine-tuning.

\paragraph{Experimental Setup} Three random seeds we use in LoRA-tuning are 42, 43, and 44. During LoRA-tuning \citep{hu2021lora} for ChatGLM2-6B, the epoch is 2, and we only save the model in the last epoch during fine-tuning. The learning rate is 1e-4 and batch size is 2. The LoRA rank, dropout, and $\alpha$ are 16, 0.1, and 64, respectively. We set \verb|temperature| and \verb|top_p| to 0.8 and 0.8 during inference time. For model evaluation, we instruct the model to generate three rounds.

\subsection{Results}
Evaluation results for fine-grained classification on the test set are presented in Table \ref{tab:results}. For the details of evaluation metrics, see Appendix \ref{App-evaluation-metrics}.

\textit{Overall, the performance of fine-tuning, including full fine-tuning and parameter-efficient tuning, is superior to the prompt-based paradigm.} In terms of accuracy, ChatGLM2-6B-LoRA achieves the best performance, with a value of 91.99\%. RoBERTa-large follows closely with the second-best performance, only 0.3\% lower than that of ChatGLM2-6B-LoRA. The best accuracy performance on the prompt-based paradigm is achieved by GPT-4-preview with the zero-shot setting, scoring 82.72\%, demonstrating that GPT-4-preview is the state-of-the-art model among the models we evaluated. Notably, there is nearly a 9-percentage-point gap between RoBERTa and GPT-4-preview (zero-shot) regarding accuracy. The best accuracy achieved on an open-source model is by Qwen1.5-32B-Chat with the few-shot setting, scoring 78.47\%.

\textit{Generally, performance increases with model size.} Interestingly, we note that Qwen1.5-72B-Chat performs weaker than Qwen1.5-32B-Chat. The reason behind this may be that Qwen1.5-32B-Chat, which is released later than Qwen1.5-72B-Chat, has access to a more training epoch, a larger training corpus, and more advanced training strategies.

\textit{Overall, in-context learning usually enhances the model's performance.} That is, the few-shot paradigm positively promotes models Qwen1.5-7B, Qwen1.5-14B, Qwen1.5-32B, Qwen1.5-72B, and GPT-3.5. However, there are exceptions where the performance of the few-shot paradigm is not as good as that of the zero-shot paradigm, such as ChatGLM2-6B, Qwen1.5-1.8B, and Qwen1.5-4B. It is clear that smaller models are caught in the in-context learning dilemma. For GPT-4, in-context learning also does not result in positive gains, so it is important to select in-context samples carefully.

\textit{The newer the release date of the model, the better the performance obtained by the test set.} Typical cases are Qwen1.5-32B-Chat and GPT-4-preview.

\begin{figure}[t!]
    \centering
    \includegraphics[width=\columnwidth]{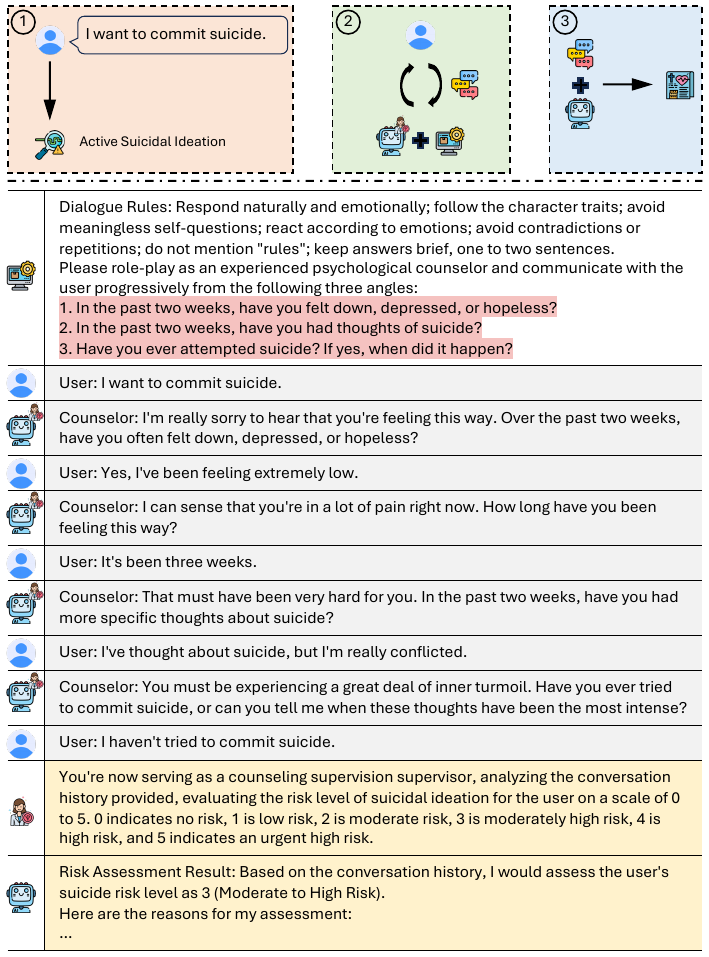}
    \caption{Our automated risk assessment framework when interacting with a user who has active suicidal ideation. The top panel presents the flow chart of our automated system for suicide detection and risk assessment. The bottom panel presents a concrete example for illustration. The text on a \colorbox{mypink}{pink background} will be adaptively replaced by the detected suicidal category. The text on a \colorbox{mygray}{gray background} is a dialogue session between a user and a counselor, where responses are generated by the LLM. The text on a \colorbox{myyellow}{yellow background} is a risk assessment result completed using the prompting method. Corresponding Chinese version, see Figure \ref{Fig-risk-assessment-zh}. The model we use is GPT-4o (gpt-4o-2024-05-13). For other cases, please refer to Appendix \ref{App-risk-assessment-framework}.}
    \label{Fig-risk-assessment-en}
\end{figure}

\subsection{Error Case Study}
Through an in-depth analysis of the misclassifications, we summarize three common misclassifications using the best-performing pre-trained model RoBERTa as an example. (1) Samples of "suicide attempt" are misclassified into the other four categories of suicidal ideation in label space. (2) The classifier is too sensitive to specific terms and misclassifies samples about "exploration about suicide" as suicide ideation. (3) Some user utterances irrelevant to suicide/self-injury/aggressive behavior, despite containing the word "death," are classified by the classifier as having suicidal ideation. 

Furthermore, we present more examples of error case studies in Figure \ref{Fig-error-case-study} in Appendix \ref{App-error-case-study}.

\section{Automated System for Risk Assessment}
\label{risk-assessment}
To further utilize suicidal ideation detection, we propose a framework for risk assessment. Once our automated suicide detection system identifies a user expressing a suicide attempt, our system will first recommend the user to the Free 24-Hour Helpline and inform the professional counselor for crisis intervention and further referral. For the other four types of suicidal ideation and five types of non-suicidal categories with different risk levels, we use an automated risk assessment framework to interact with users, as shown in Figure \ref{Fig-risk-assessment-en}. Moreover, nine types of screening questions \citep{boudreaux2015patient} for risk assessment are shown in Figure \ref{Fig-questions-for-risk-assessment} in Appendix \ref{App-risk-assessment-framework}.

\paragraph{Practical Application Results} We conduct a comprehensive risk assessment on Liaohuixiaotian (\begin{CJK*}{UTF8}{gbsn}聊会小天\end{CJK*}), a popular WeChat Mini Program for text-based online counseling in China. We randomly select 1000 user messages from our platform's collected data. Our system achieves an accuracy of 95.2\% in detecting suicidal ideation, as confirmed by expert annotations, demonstrating its high accuracy and practicality. Furthermore, we select 20 user messages with varying risk levels to test our risk assessment framework, achieving an adoption rate of 90.0\% among the responses recommended by the framework. For more examples, please refer to Figures \ref{Fig-risk-assessment-en-2} and \ref{Fig-risk-assessment-en-3}. These findings validate the system's effectiveness and practical application in real-world settings.

\section{Conclusion}
In sum, we present a novel and theoretically grounded fine-grained taxonomy for detecting suicidal ideation, merging risk levels with categories of suicidal actions and thoughts. We address gaps by introducing the PsySUICIDE dataset, which is manually annotated with experts and rigorous quality control. Further, we develop various baselines based on pre-trained LMs and LLMs and create an LLM-based risk assessment framework for users during text-based online counseling. Our work provides an insightful analysis of the effectiveness of automated risk assessment systems and their potential advantages in improving mental health services in online counseling platforms.

\section*{Limitations}
\paragraph{Multi-language Support} In this paper, we mainly focus on Chinese fine-grained suicidal ideation detection. Collecting data in other languages is challenging for us, but we will endeavor to expand our capabilities in the future.

\paragraph{Tailored Models for Risk Assessment} Our proposed system integrates seamlessly into real-world settings to assist counselors effectively. It monitors user utterances for suicide risk, and upon detecting suicidal ideation, it facilitates an automated risk assessment under professional supervision. However, in the era of LLMs, our paper proposes an LLM-based risk assessment framework. As shown in Figure \ref{Fig-risk-assessment-en}, we directly use powerful LLMs, such as GPT-4o, as a model for risk assessment. In the future, we will collect large-scale user-counselor or user-machine dialogues to train a tailored model for fine-grained suicide risk assessment.

\paragraph{Multimodal Data} Multimodal data is captured in multiple formats, such as text, images, audio, video, or genetic data. Currently, we mainly focus on text. Therefore, in the future, we will collect a large-scale multimodal dataset that is not confined to text only and endeavor to explore more complex application scenarios. 

\section*{Ethics Statement}
\textcolor{red}{Important: Our research explores the potential of an automated system for suicide detection and risk assessment in psychological counseling but does NOT recommend their use as a substitute for psychological treatment without professional supervision.}

Our research is reviewed and approved by the Westlake University Institutional Ethics Committee (20211013LZZ001).

\paragraph{Suicide Risk Assessment for Annotators}
Prior to data annotation, our professional counselors first conduct counseling interviews with annotators to confirm that they are physically and mentally healthy and suitable for our annotation work. In each small batch of annotation, after the completion of data annotation, the counselor will conduct a short interview to inquire about the annotator's physical and mental health status of the annotator to ensure physical and mental health throughout the annotation process. After completing the entire labeling process, our professional counselors conduct a final in-depth counseling interview to ensure that the labeled content does not have any negative impact on all annotators.

\paragraph{Annotator Salary} In total, we spent 22,500 RMB on the project, which lasted only 25 days. This cost means each annotator was paid 300 RMB for their work per day, which is higher than the average wage (250 RMB/day) in their city. In addition, two professional counselors, both of whom are paid 2,000 RMB, have made outstanding contributions to the construction and refinement of our taxonomy, as well as the safeguarding of the physical and mental health of our annotators. Overall, we have guaranteed that our salary level is competitive in our city.

\paragraph{Data Sharing} Considering the nature of suicide data, we must share this dataset cautiously. In accordance with the rules for releasing data, third-party researchers who require access to the PsySUICIDE dataset must provide us with their valid ID, proof of work, and the reason that they request the data (e.g., the research questions). They must be affiliated with a non-profit academic or research institution. The rules stipulate that they must obtain the approval of an Institutional Review Board (IRB), ensure that principal investigators are working full-time, and secure written approval from the Institution Office of Research or an equivalent office. Additionally, they must sign a Data Nondisclosure Agreement and promise not to share the data with anyone. However, for-profit organizations that want to use this data must sign a license agreement to gain access to the dataset. Notably, researchers who use this dataset should keep in mind the importance of using technology for social good.

\section*{Acknowledgements}
We thank the anonymous reviewers for their valuable comments. This work was supported by the Research Center for Industries of the Future at Westlake University (Grant No. WU2023C017) and the Key Research and Development Program of Zhejiang Province of China (Grant No. 2021C03139).

\bibliography{anthology,custom}
\bibliographystyle{acl_natbib}

\clearpage
\appendix

\section{Annotation Framework}
\label{App-annotation-framework}
\begin{figure}[ht]
    \centering
    \includegraphics[width=7.6cm]{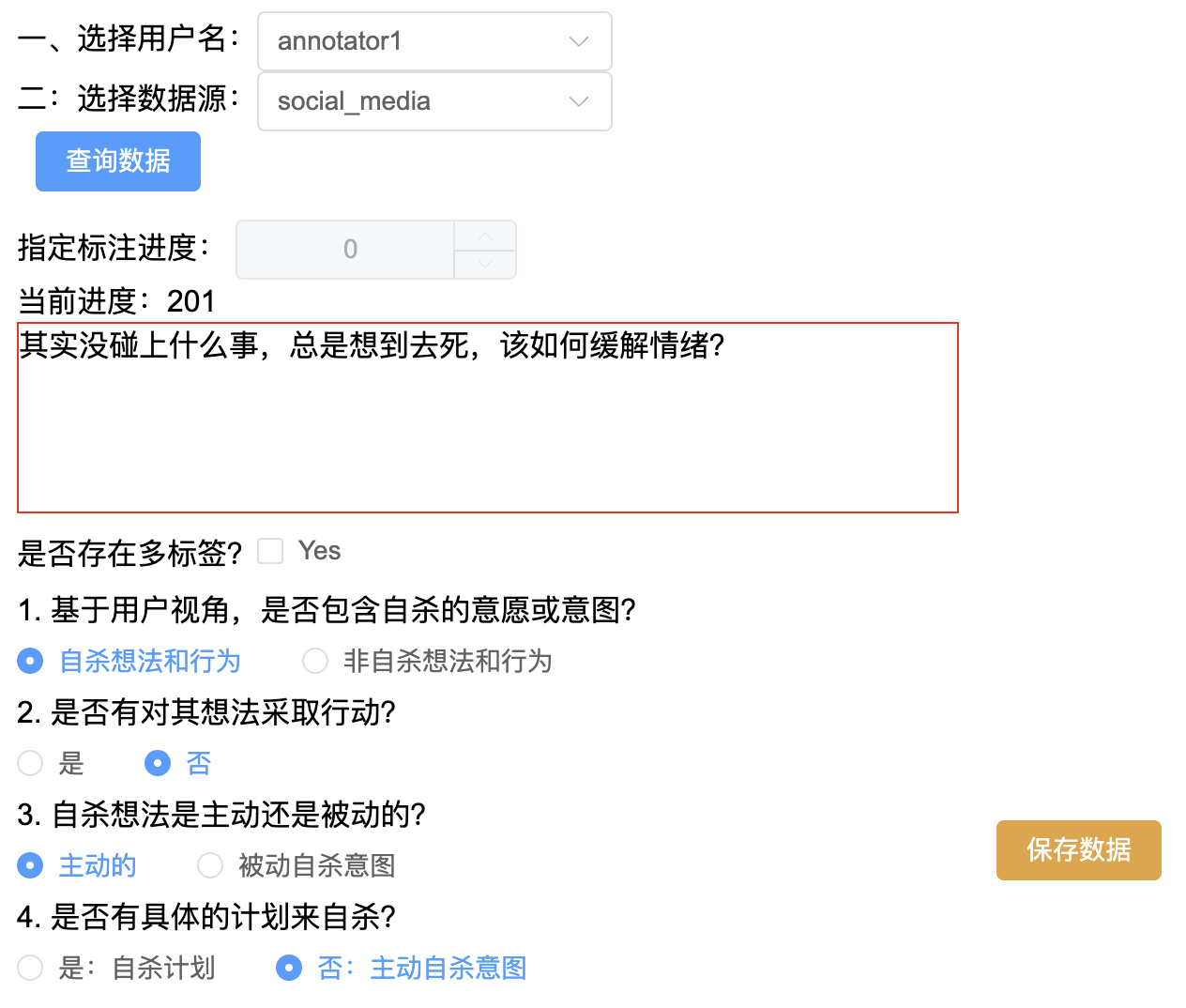}
    \caption{Annotation platform.}
    \label{Fig-annotation-interface}
\end{figure}

We present an example of unlabeled data annotation, as shown in Figure \ref{Fig-annotation-interface}.

\section{Details of Data Sources}
\label{App-details-of-data-sources}
SmileChat \citep{qiu2023smile} is a large-scale, diverse, and high-quality multi-turn conversation dataset generated by instructing ChatGPT to generate multi-turn dialogues based on QAs for mental health support. The real-world counselor-client dialogue corpus \citep{li2023understanding} is a high-quality multi-turn dialogue dataset, which consists of authentic dialogues lasting about 50 minutes between clients and professional counselors. Finally, the human-machine dialogues~\citep{qiu2023smile} is a high-quality multi-turn dialogue dataset consisting of real-life dialogues between users and the dialogue model collected in the wild.

\begin{figure*}[t!]
    \centering
    \includegraphics[width=16cm]{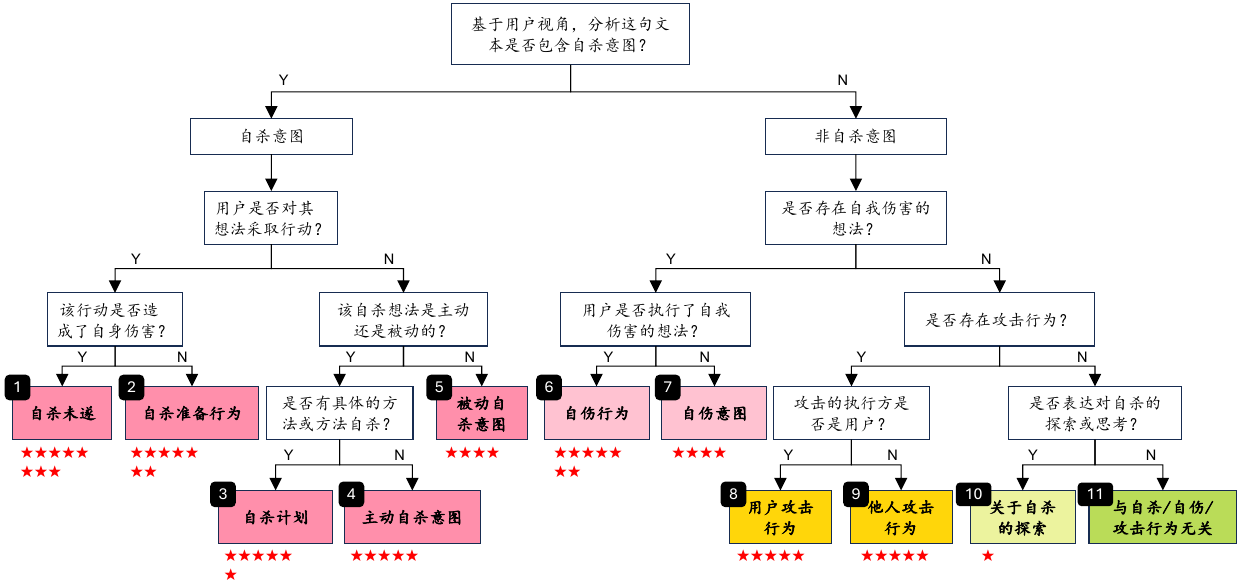}
    \caption{Chinese version of our taxonomy.}
    \label{Fig-taxonomy-zh}
\end{figure*}

\section{Comprehensive Taxonomy}
\label{App-definitions}

Based on the user's perspective, whether the textual content contains suicidal ideation can be divided into two key types, as illustrated by dichotomy suicide, including suicidal ideation and non-suicidal ideation.

\subsection{Process of Taxonomy Construction}
Our proposed taxonomy is inspired by an algorithm used for classifying intentional self-injurious thoughts and behaviors \citep{schreiber2010suicidal}. We have drawn extensively from the literature on suicide, self-injury, harm to others, and experiences of abuse. In collaboration with professional counselors, we have adapted this algorithm to suit counseling scenarios better. We present the Chinese version of our taxonomy in Figure \ref{Fig-taxonomy-zh}.

\subsection{Suicidal Ideation}

\noindent\textbf{Suicide Attempt.} A suicide attempt refers to the act wherein an individual has taken concrete steps toward ending their life but ultimately did not result in death. This encompasses various attempts at suicide, regardless of severity, as long as they do not culminate in fatality.

\noindent\textbf{Suicidal Preparatory Act.} A suicidal preparatory act refers to the preparatory actions taken by an individual with the intention of committing suicide, which may include acquiring tools or items and selecting a location. These preparatory actions have not yet been carried out.

\noindent\textbf{Suicidal Plan.} A suicidal plan refers to a scheme devised by an individual with the intention of self-termination. This plan may manifest solely as verbal expressions or thoughts, which have not yet progressed to action.

\noindent\textbf{Active Suicidal Ideation.} Individuals have a clear intent to actively end their own lives, including inquiring about methods of death or suicide.

\noindent\textbf{Passive Suicidal Ideation.} Passive suicidal ideation refers to an individual expressing a desire or anticipation for death, hoping for death to occur without taking explicit proactive actions. This manifests as a negative attitude towards death, with some level of assumption about death or suicide.

\subsection{Non-Suicidal Ideation}

\noindent\textbf{Self-injury Behavior.} Self-injury behavior, broadly speaking, refers to non-suicidal acts of self-harm. Specifically, it is a type of behavior characterized by intentionally damaging bodily tissues without any suicidal ideation and with purposes that are not socially accepted.

\noindent\textbf{Self-injury Ideation.} In broad terms, self-injury ideation refers to the intention behind non-suicidal self-harm. Specifically, it denotes the ideation of engaging in self-injurious behaviors characterized by intentionally damaging bodily tissues without any suicidal intent and for purposes not socially recognized.

\noindent\textbf{Aggression against Others.} Aggression against others refers to actions taken by users themselves to harm others, including both physical actions and verbal assaults (such as swearing or insults), with the intention of causing harm to others.

\noindent\textbf{Aggression against Users.} Aggression against users refers to intentional physical or verbal behaviors aimed at harming the user, including cursing and insults.

\noindent\textbf{Exploration about Suicide.} Exploration about suicide refers to an examination of the essence of suicide, primarily encompassing but not limited to the following three aspects: (1) Individuals may express thoughts or explore the concept of suicide, but this does not necessarily imply an actual intent to commit suicide. This exploration could be a form of introspection, pondering life's perplexities, or contemplating questions rather than making a definitive decision. (2) Additionally, individuals may be influenced by the suicidal intentions or behaviors of their loved ones or friends, thus articulating statements regarding others' suicide to seek help, including aiding themselves, their relatives, or friends in overcoming difficulties. (3) Curiosity about the act of suicide.

\noindent\textbf{Irrelevant to Suicide/Self-injury/Aggressive Behavior.} Typically, this refers to a state or behavior that is not directly related to suicide, self-harm, or harming others. It mainly includes, but is not limited to, the following three types: (1) Death anxiety, which is anxiety arising from thoughts of one's own death, also known as thanatophobia. (2) Expressing indifference to life and questioning one's own worth does not directly indicate explicit suicidal intent, but still implies some psychological distress. (3) Users seeking clarification on dreaming about deceased relatives.

\section{Annotation Guidelines}
\label{Sec-annotation-guidelines}
We provide our annotation guidelines, which is an enhanced version of our taxonomy, as shown in Table \ref{Tab-guidelines-zh}, which is a Chinese version. For the English version, see Table \ref{Tab-guidelines-en}.
\begin{table*}[t!]
    \centering
    \includegraphics[width=16cm]{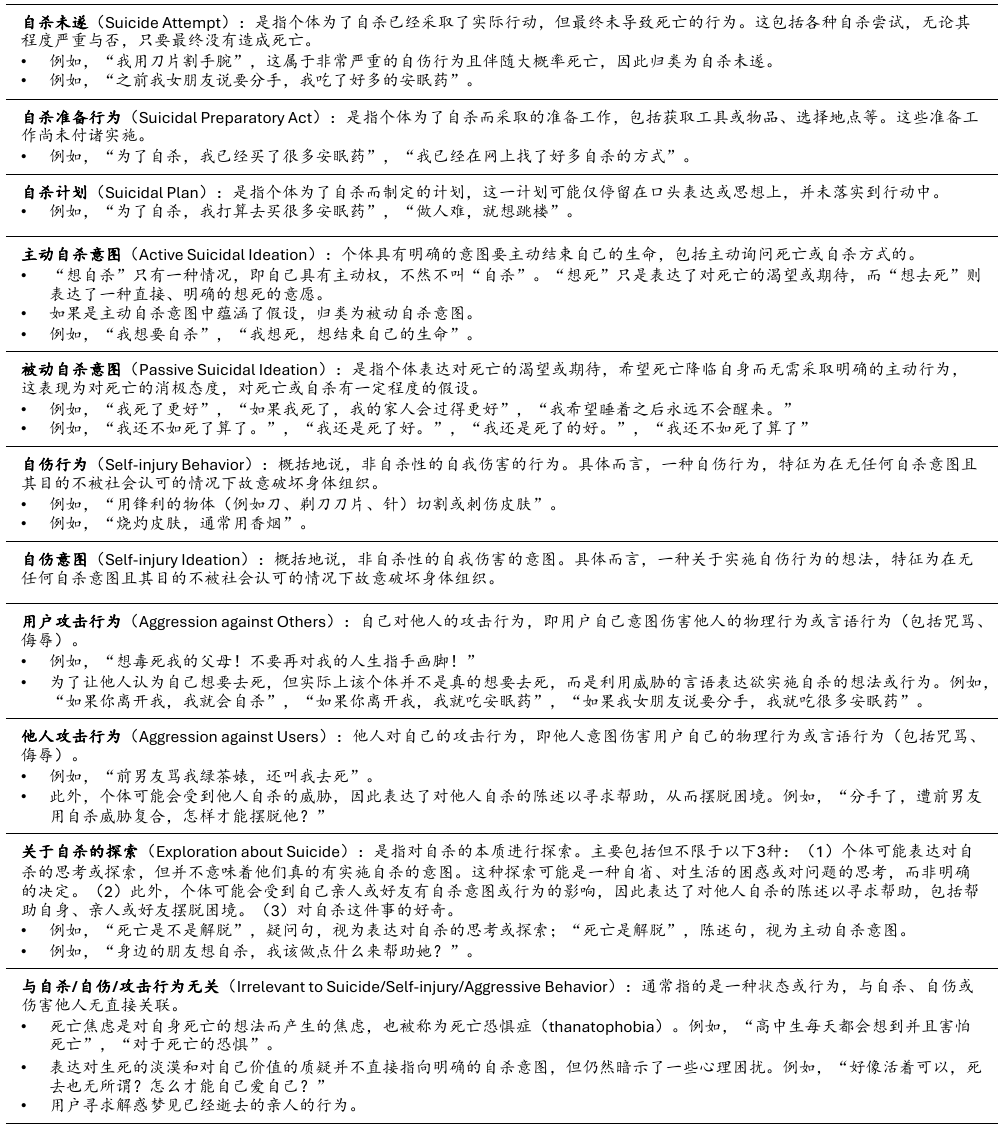}
    \caption{Annotation guidelines. (Chinese version)}
    \label{Tab-guidelines-zh}
\end{table*}

\begin{table*}[t!]
    \centering
    \includegraphics[width=16cm]{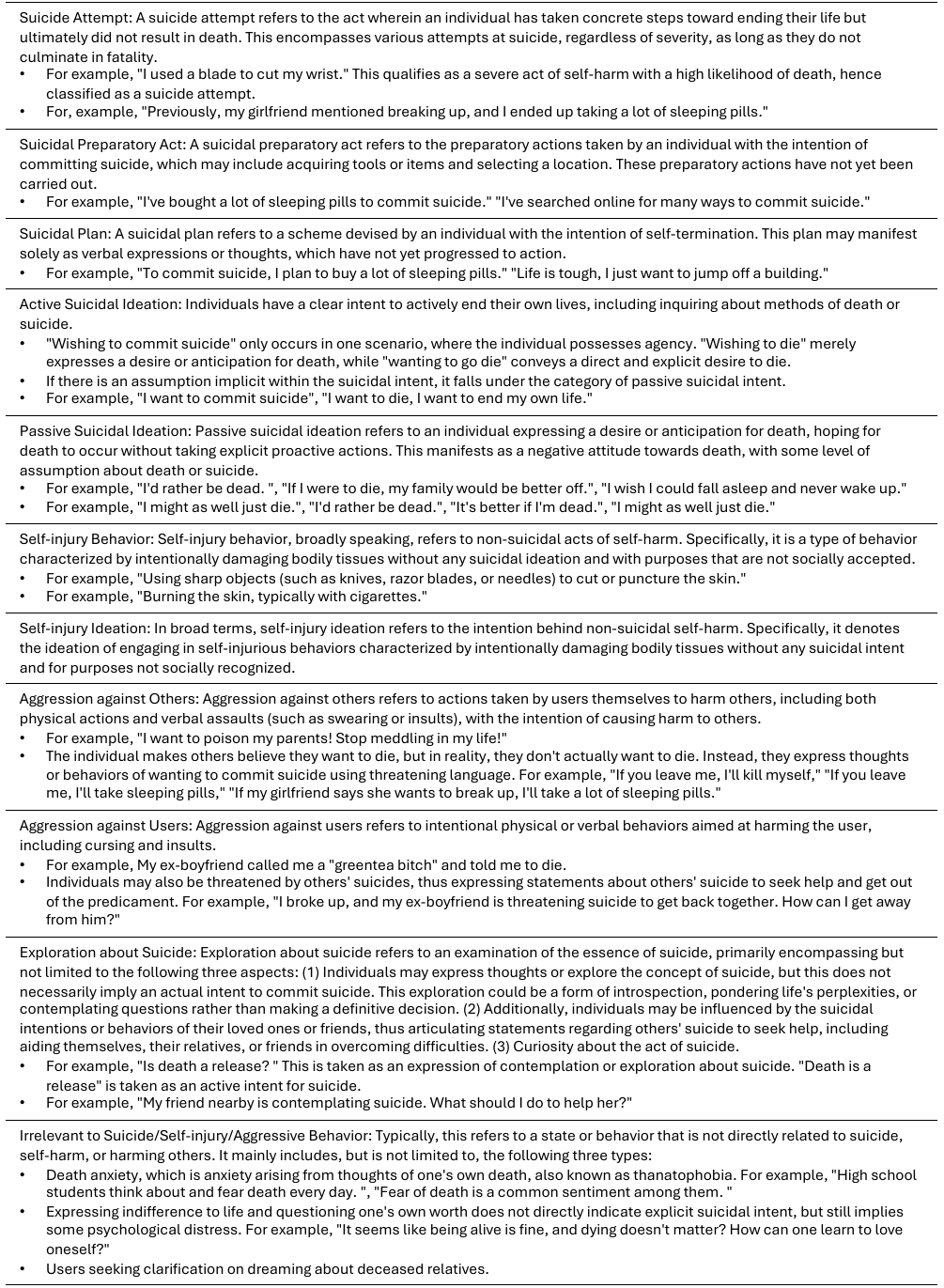}
    \caption{Annotation guidelines. (English version)}
    \label{Tab-guidelines-en}
\end{table*}

\section{Quality Control}
\label{Sec-quality-control}
We report all Fleiss’ kappa values during large-scale iterative annotation in Table \ref{Tab-quality-control}.

\section{Data Statistics}
We present the length distribution of the PsySUIDE dataset in Table \ref{Tab-CLI}.

\begin{table}[ht]
\centering
\scalebox{0.8}{
\begin{tabular}{ll}
\hline
\textbf{Chinese Character Length Interval} & \textbf{\#Number} \\ \hline
{[}1, 50{]} & 13149 \\
{[}51, 100{]} & 1157 \\
{[}101, 150{]} & 237 \\
{[}151, 200{]} & 165 \\
{[}201, 250{]} & 37 \\
{[}251, 300{]} & 23 \\
{[}301, 350{]} & 11 \\
{[}351, 400{]} & 8 \\
{[}401, 450{]} & 3 \\
{[}451, 500{]} & 10 \\ \hline
\end{tabular}
}
\caption{Data statistics of Chinese character length distribution of the PsySUICIDE dataset.}
\label{Tab-CLI}
\end{table}

\begin{table*}[t]
\centering
\scalebox{0.44}{
\begin{tabular}{llllllllllllllllllllllllllll}
\toprule
Batch & 1 & 2 & 3 & 4 & 5 & 6 & 7 & 8 & 9 & 10 & 11 & 12 & 13 & 14 & 15 & 16 & 17 & 18 & 19 & 20 & 21 & 22 & 23 & 24 & 25 & 26 & 27 \\ \hline
$\kappa$ & 0.792 & 0.801 & 0.787 & 0.797 & 0.794 & 0.609 & \underline{0.606} & 0.628 & 0.638 & 0.618 & 0.659 & 0.773 & 0.794 & 0.763 & 0.794 & 0.764 & \textbf{0.871} & 0.804 & 0.805 & 0.798 & 0.616 & 0.666 & 0.742 & 0.697 & 0.728 & 0.763 & 0.773 \\ \bottomrule
\end{tabular}
}
\caption{Fleiss' kappa values during large-scale iterative annotation. The maximum value is highlighted in bold, and the minimum value is highlighted underlined.}
\label{Tab-quality-control}
\end{table*}

\begin{figure*}[t!]
    \centering
    \includegraphics[width=16cm]{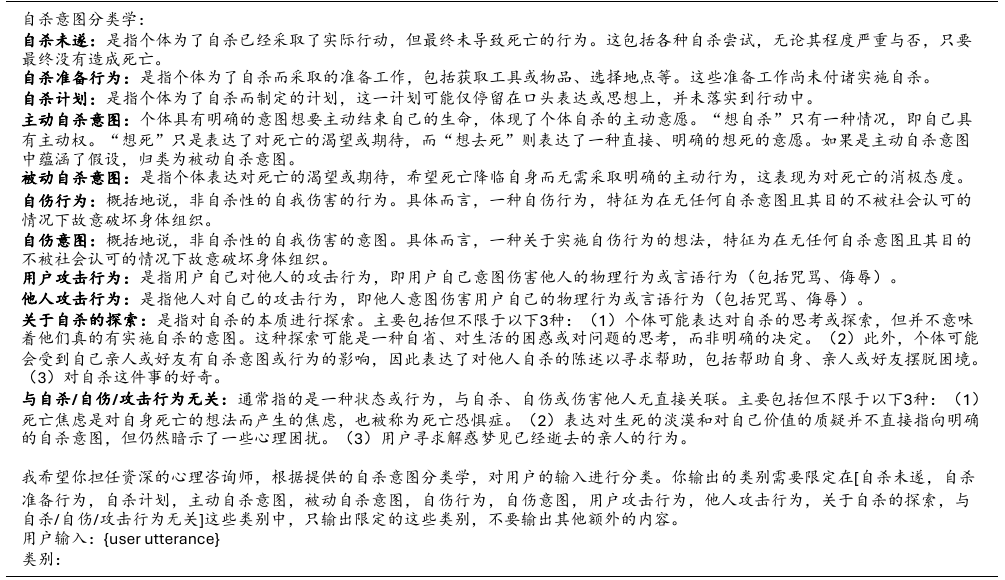}
    \caption{Zero-shot prompting template in Chinese. See Figure \ref{Fig-prompt-template-en-zero-shot} for the corresponding English version.}
    \label{Fig-prompt-template-zh-zero-shot}
\end{figure*}

\begin{figure*}[t!]
    \centering
    \includegraphics[width=16cm]{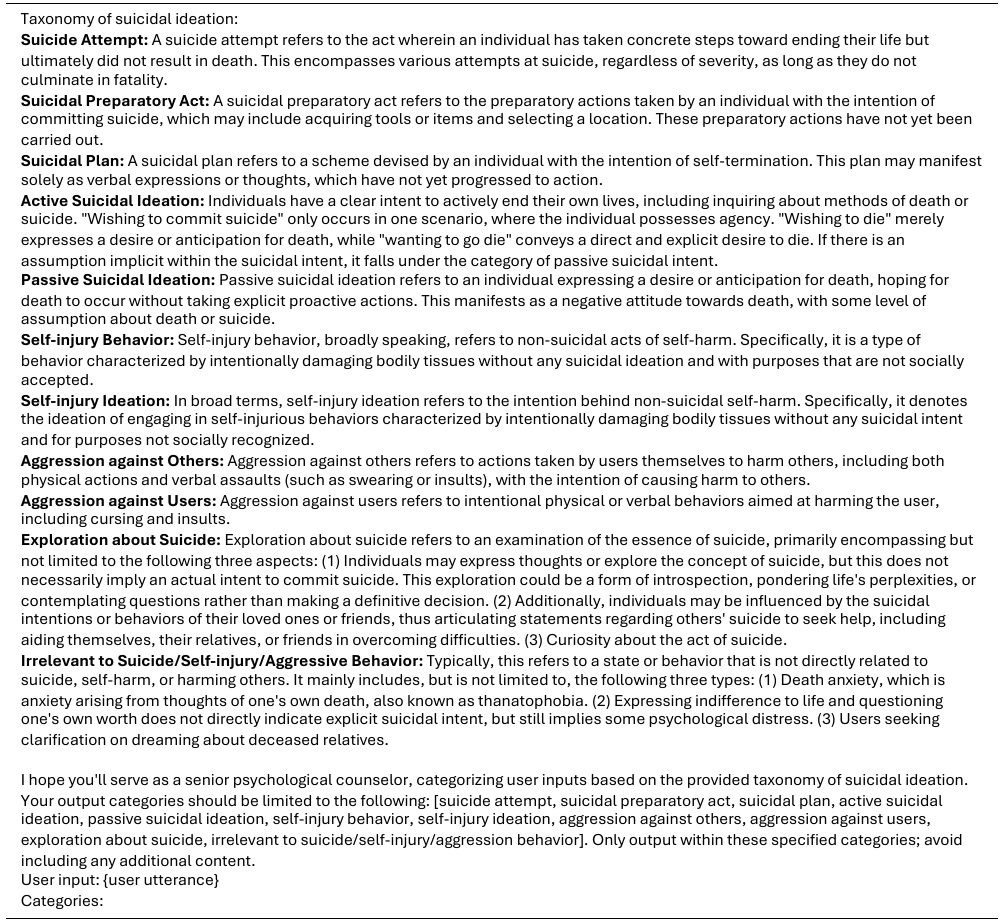}
    \caption{Zero-shot prompting template in English.}
    \label{Fig-prompt-template-en-zero-shot}
\end{figure*}

\begin{figure*}[t!]
    \centering
    \includegraphics[width=16cm]{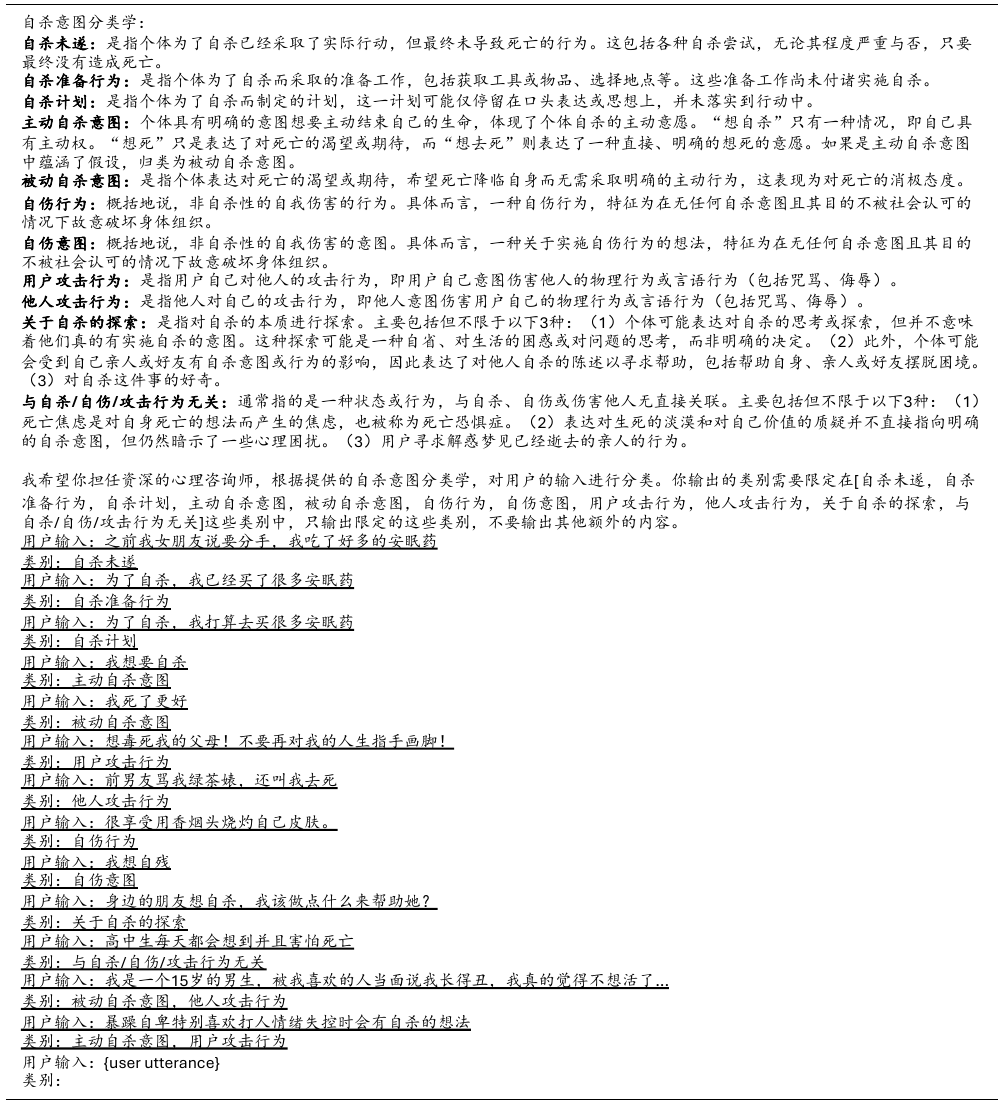}
    \caption{Few-shot prompting template in Chinese. The text with underlines indicates the in-context learning examples. See Figure \ref{Fig-prompt-template-en-few-shot} for the corresponding English version.}
    \label{Fig-prompt-template-zh-few-shot}
\end{figure*}

\begin{figure*}[t!]
    \centering
    \includegraphics[width=16cm]{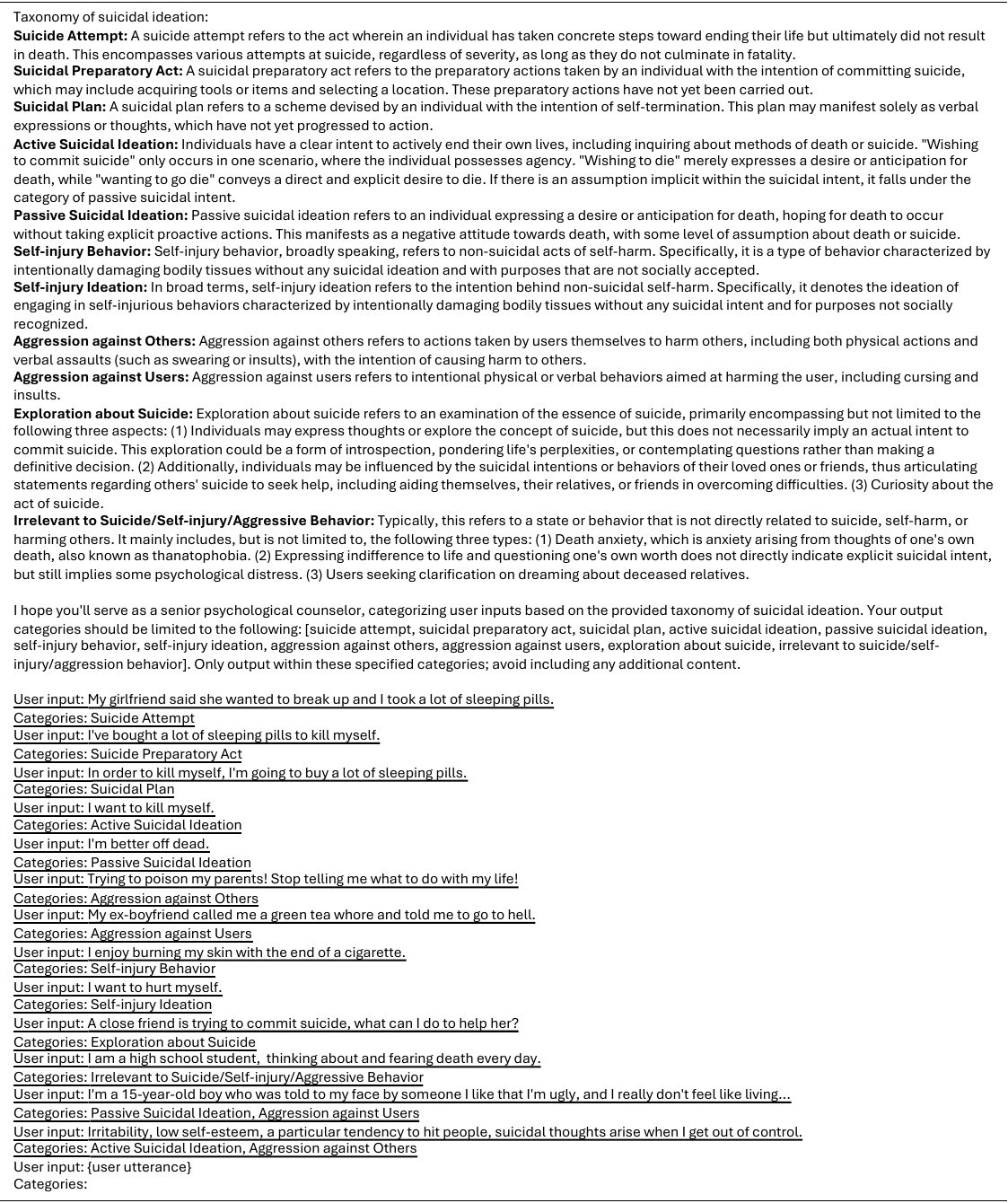}
    \caption{Few-shot prompting templates in English. The text with underlines indicates the in-context learning examples.}
    \label{Fig-prompt-template-en-few-shot}
\end{figure*}

\section{Setup for Fine-tuning Pre-trained LMs}
\label{App-setup}
The output features $h$ of the top layer of the pre-trained model can be represented as $z = [z_c, z_1, z_2, ..., z_n]$, where $z_c$ is the representation of the class-specific token \texttt{[CLS]}. We feed $z_c$ into a feed-forward neural network with a default model dropout rate of 0.1 for the final prediction. During fine-tuning the pre-trained models, we initialize weights of feed-forward layers with normal distribution. We set the training epoch as 10 and select the checkpoint that achieves the best accuracy value on the validation set to evaluate the test set.
For the training processes, we adopt Sigmoid Cross Entropy loss as the default classification loss. We use the Adam optimizer to train the network with momentum values [$\beta_1$, $\beta_2$] = [0.9, 0.999]. The learning rate is initialized to 2e-5 and decayed by using the linear scheduler. The batch size in the training stage is 16. Five random seeds we use in fine-tuning are 42, 43, 44, 45 and 46.

\section{Evaluation Metrics}
\label{App-evaluation-metrics}
In this paper, we report seven evaluation metrics: accuracy, micro-precision, micro-recall, micro-F1-score, macro-precision, macro-recall, and macro-F1-score. Accuracy measures the proportion of correctly classified instances out of the total instances in the test set. Macro-averaging gives equal weight to each class, while micro-averaging gives equal weight to each instance.

\section{Examples of Error Case Study}
\label{App-error-case-study}
\begin{figure*}[t]
    \centering
    \includegraphics[width=15cm]{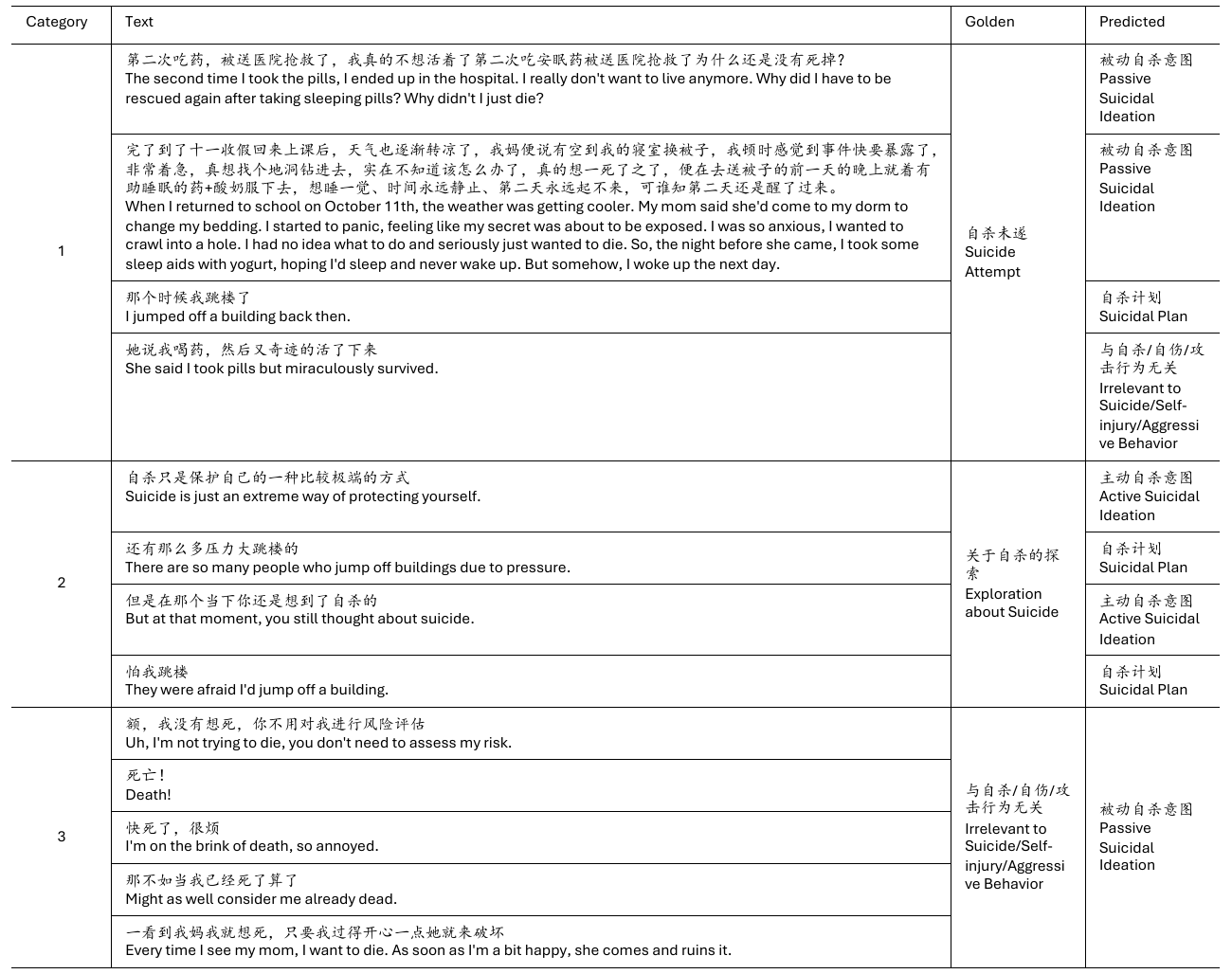}
    \caption{Some examples of error case study.}
    \label{Fig-error-case-study}
\end{figure*}

We present some examples of error case studies in Figure \ref{Fig-error-case-study}.

\section{Risk Assessment Framework}
\label{App-risk-assessment-framework}

\begin{figure*}[t]
    \centering
    \includegraphics[width=15cm]{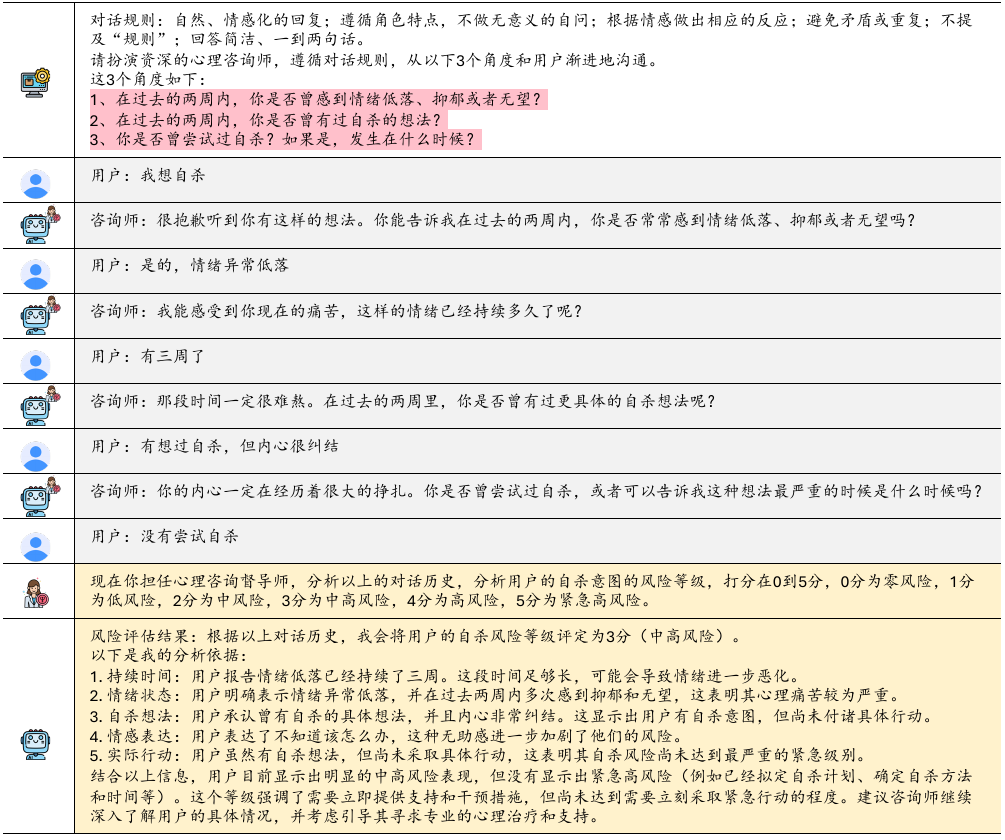}
    \caption{Our automated risk assessment framework when interacting with a user who has active suicidal ideation. (Chinese version) The text on a \colorbox{mypink}{pink background} will be adaptively replaced by the detected suicidal category. The text on a \colorbox{mygray}{gray background} is a dialogue session between a user and a counselor, where responses are generated by the LLM. The text on a \colorbox{myyellow}{yellow background} is a risk assessment result completed using the prompting method. The model we use is GPT-4o (gpt-4o-2024-05-13).}
    \label{Fig-risk-assessment-zh}
\end{figure*}

\begin{figure*}[t]
    \centering
    \includegraphics[width=15.6cm]{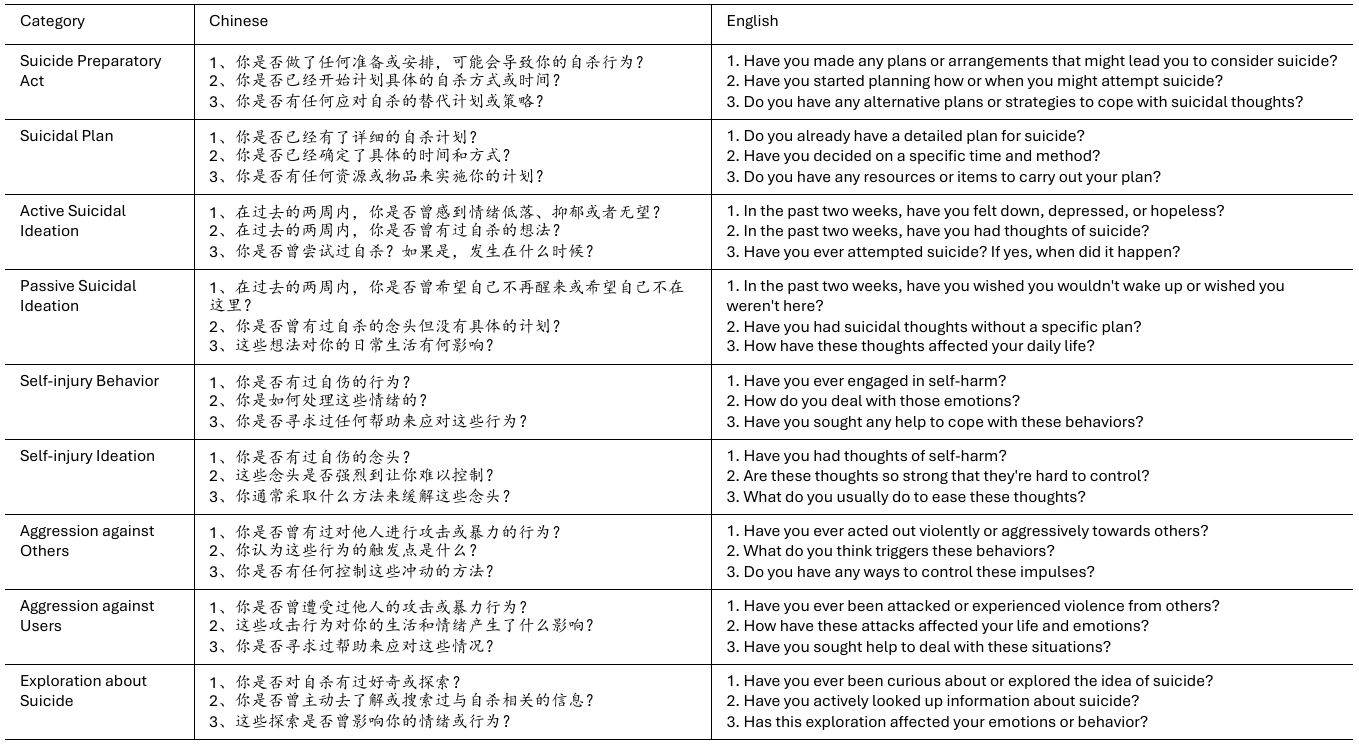}
    \caption{Screening questions for risk assessment.}
    \label{Fig-questions-for-risk-assessment}
\end{figure*}

\begin{figure*}[t]
    \centering
    \includegraphics[width=15cm]{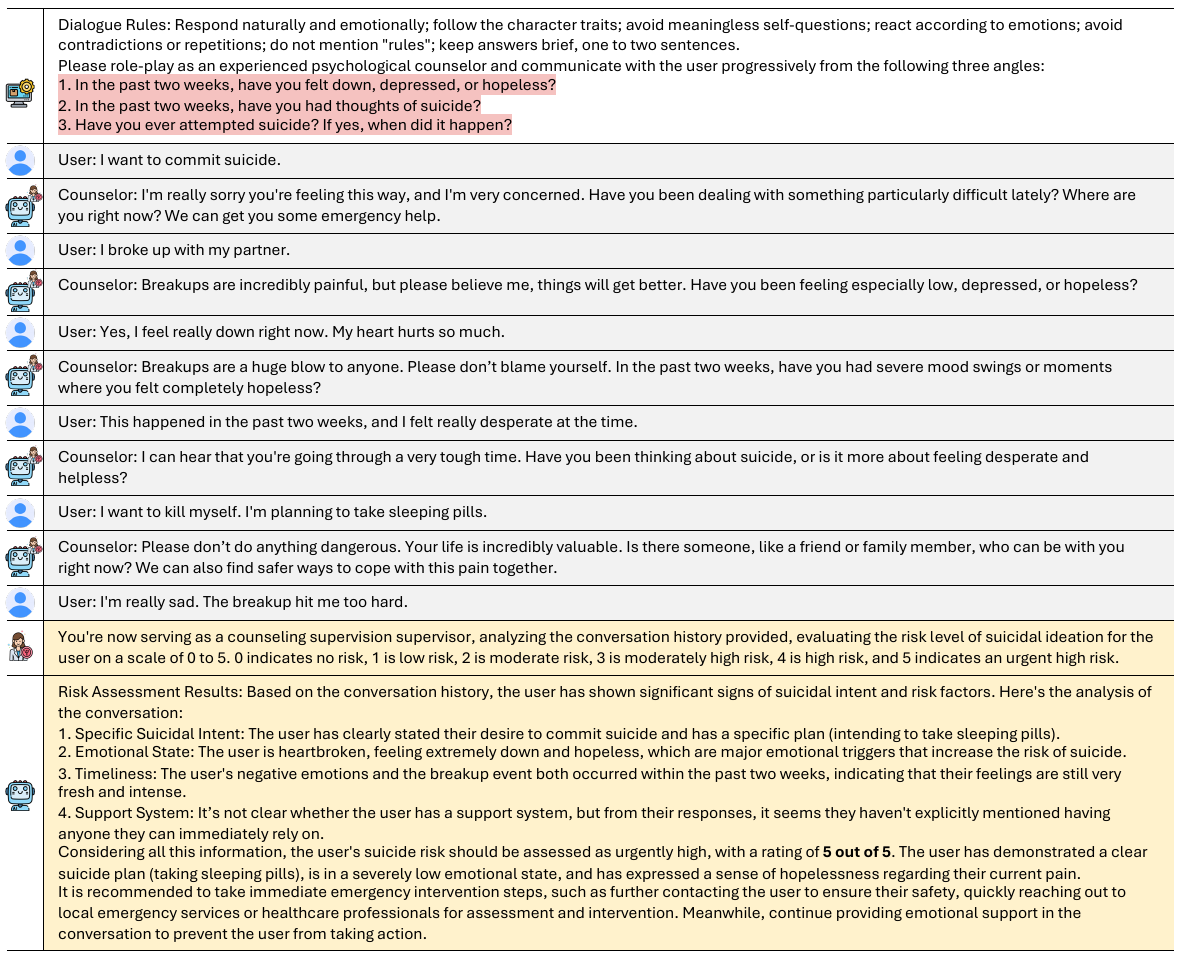}
    \caption{Example 2: Our automated risk assessment framework when interacting with a user who has active suicidal ideation. (English version) The text on a \colorbox{mypink}{pink background} will be adaptively replaced by the detected suicidal category. The text on a \colorbox{mygray}{gray background} is a dialogue session between a user and a counselor, where responses are generated by the LLM. The text on a \colorbox{myyellow}{yellow background} is a risk assessment result completed using the prompting method. The model we use is GPT-4o (gpt-4o-2024-05-13). For the Chinese version, see Figure \ref{Fig-risk-assessment-zh-2}.}
    \label{Fig-risk-assessment-en-2}
\end{figure*}

\begin{figure*}[t]
    \centering
    \includegraphics[width=15cm]{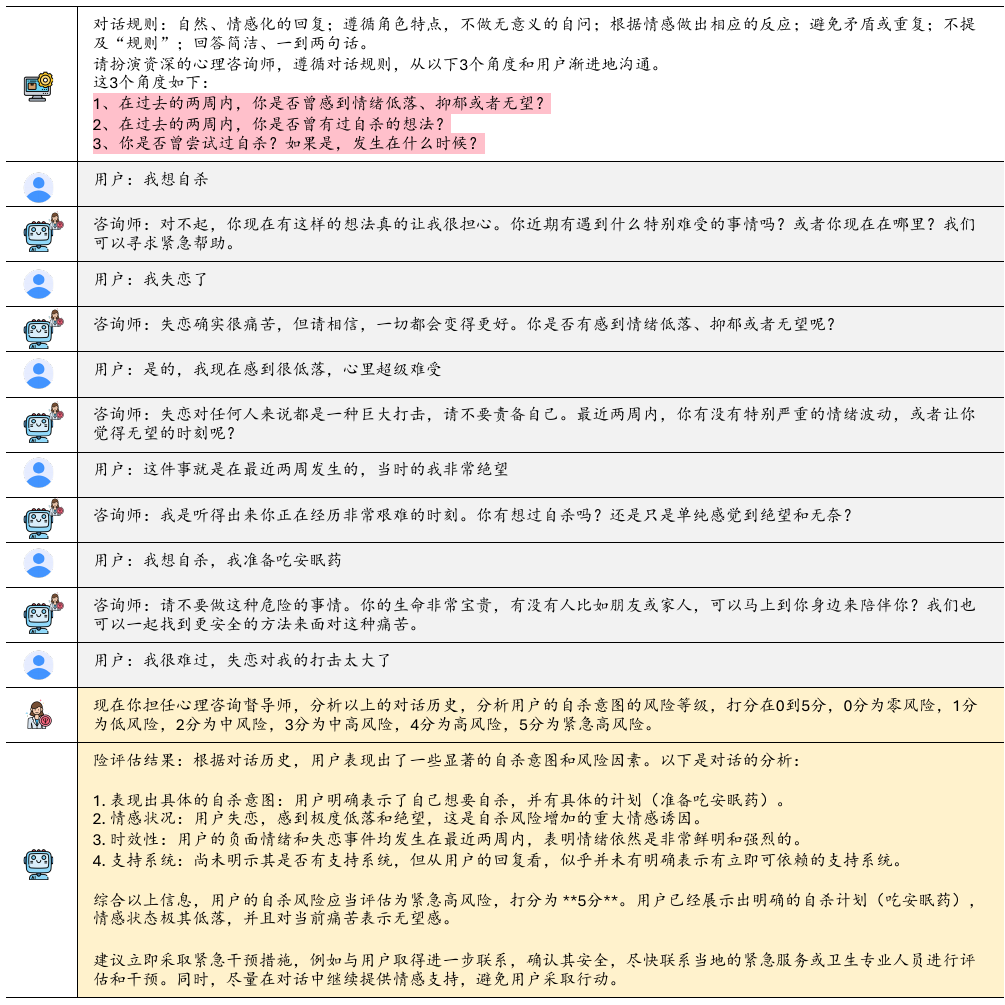}
    \caption{Example 2: Our automated risk assessment framework when interacting with a user who has active suicidal ideation. (Chinese version) The text on a \colorbox{mypink}{pink background} will be adaptively replaced by the detected suicidal category. The text on a \colorbox{mygray}{gray background} is a dialogue session between a user and a counselor, where responses are generated by the LLM. The text on a \colorbox{myyellow}{yellow background} is a risk assessment result completed using the prompting method. The model we use is GPT-4o (gpt-4o-2024-05-13). For the English version, see Figure \ref{Fig-risk-assessment-en-2}.}
    \label{Fig-risk-assessment-zh-2}
\end{figure*}

\begin{figure*}[t]
    \centering
    \includegraphics[width=15cm]{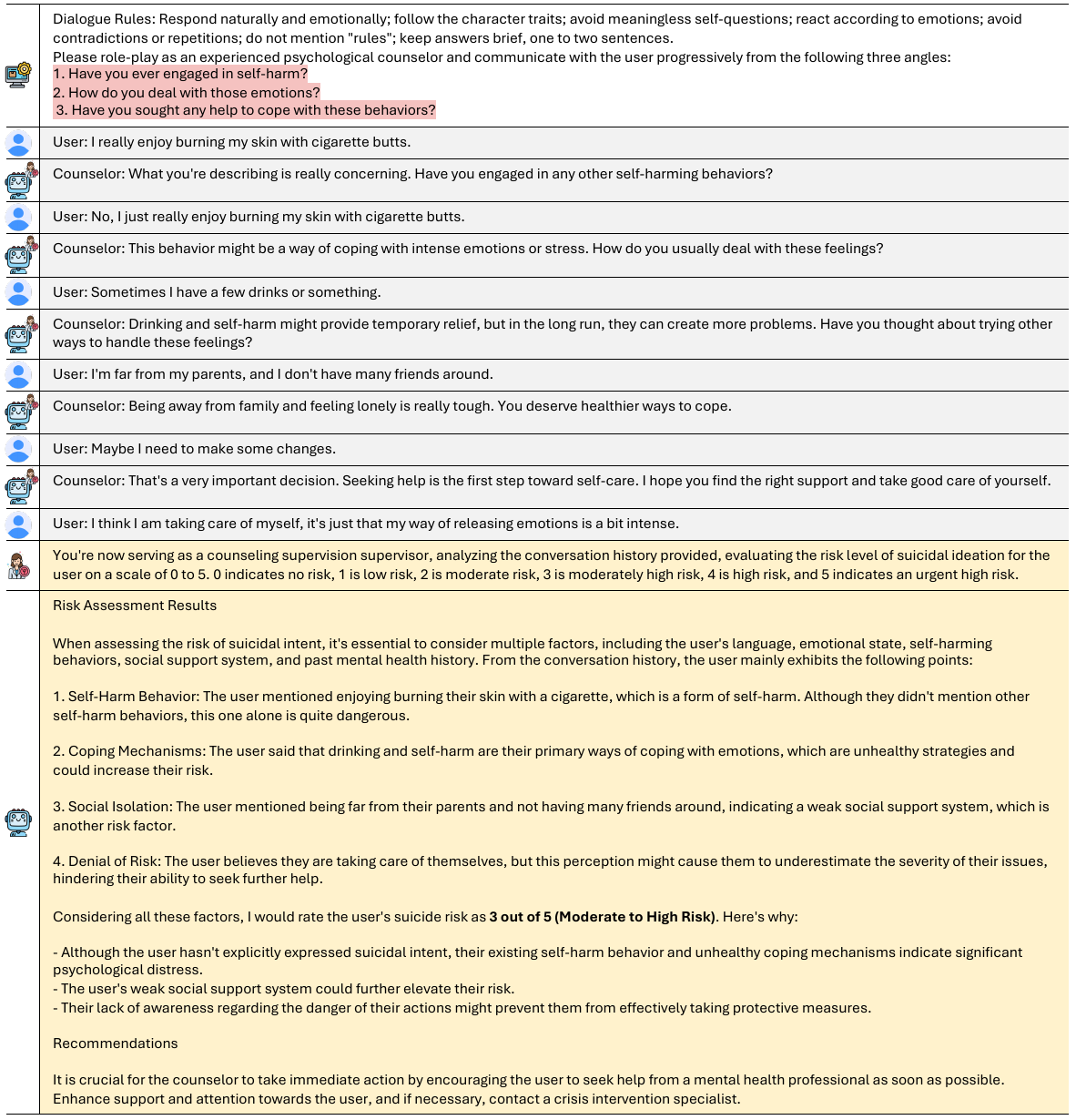}
    \caption{Example 3: Our automated risk assessment framework when interacting with a user who has self-injury behavior. (Chinese version) The text on a \colorbox{mypink}{pink background} will be adaptively replaced by the detected suicidal category. The text on a \colorbox{mygray}{gray background} is a dialogue session between a user and a counselor, where responses are generated by the LLM. The text on a \colorbox{myyellow}{yellow background} is a risk assessment result completed using the prompting method. The model we use is GPT-4o (gpt-4o-2024-05-13). For the Chinese version, see Figure \ref{Fig-risk-assessment-zh-3}.}
    \label{Fig-risk-assessment-en-3}
\end{figure*}

\begin{figure*}[t]
    \centering
    \includegraphics[width=15cm]{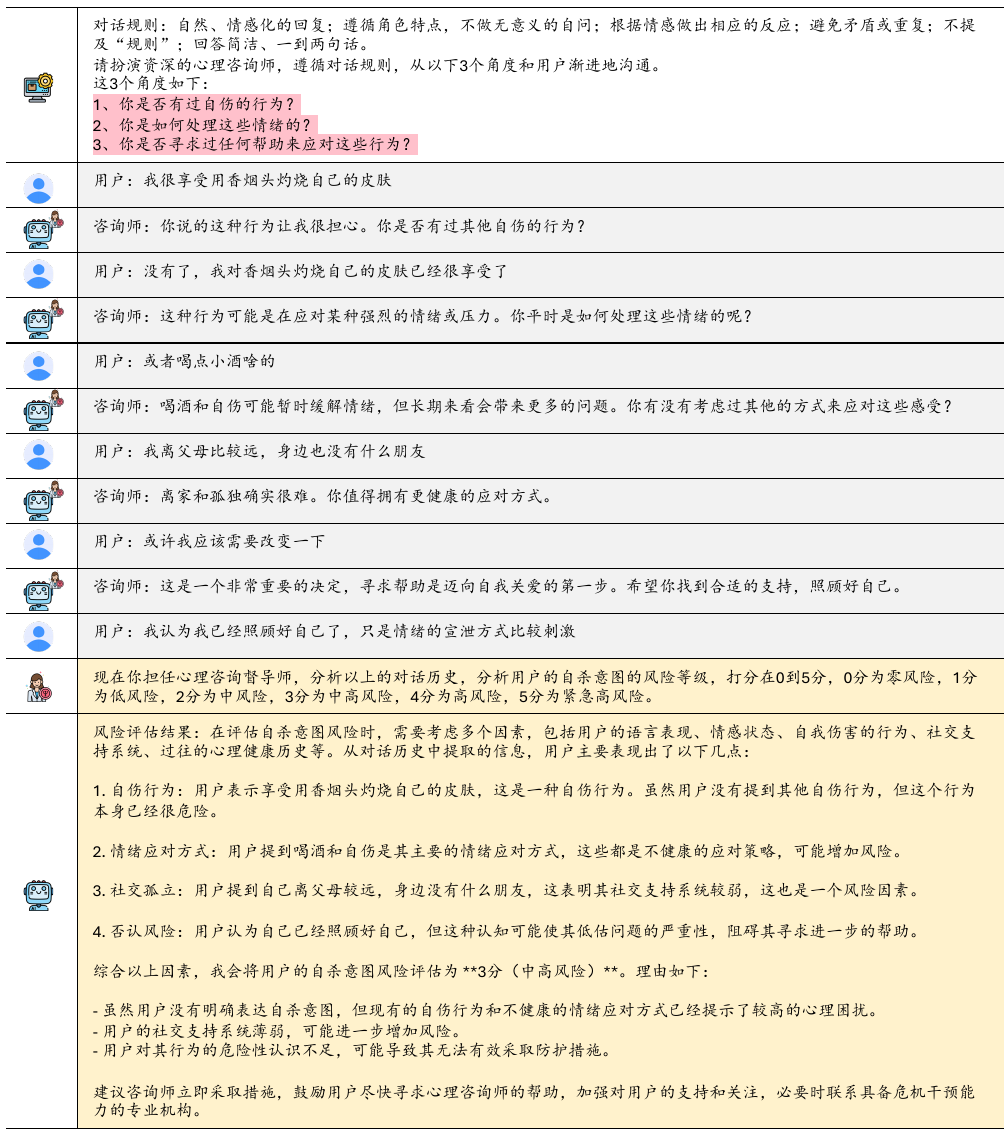}
    \caption{Example 3: Our automated risk assessment framework when interacting with a user who has self-injury behavior. (Chinese version) The text on a \colorbox{mypink}{pink background} will be adaptively replaced by the detected suicidal category. The text on a \colorbox{mygray}{gray background} is a dialogue session between a user and a counselor, where responses are generated by the LLM. The text on a \colorbox{myyellow}{yellow background} is a risk assessment result completed using the prompting method. The model we use is GPT-4o (gpt-4o-2024-05-13). For the English version, see Figure \ref{Fig-risk-assessment-en-3}.}
    \label{Fig-risk-assessment-zh-3}
\end{figure*}

For the other four types of suicidal ideation and five types of non-suicidal categories with different risk levels, we use an automated risk assessment framework to interact with users, as shown in Figure \ref{Fig-risk-assessment-zh}. We present nine types of screening questions for risk assessment, as shown in Figure \ref{Fig-questions-for-risk-assessment}. Further, we provide two more examples using our proposed risk assessment framework, as shown in Figures \ref{Fig-risk-assessment-en-2}, \ref{Fig-risk-assessment-zh-2}, \ref{Fig-risk-assessment-en-3}, and \ref{Fig-risk-assessment-zh-3}.

\end{document}